\documentclass{article}

\usepackage{microtype}
\usepackage{graphicx}
\usepackage{subcaption}
\usepackage{booktabs} 

\usepackage{float}
\usepackage{cuted}
\usepackage{caption} 

\usepackage{hyperref}

\usepackage[preprint]{icml2026}

\usepackage{amsmath}
\usepackage{amssymb}
\usepackage{mathtools}
\usepackage{amsthm}

\usepackage{booktabs}
\usepackage{graphicx}
\usepackage[table]{xcolor}   
\usepackage{array}
\usepackage{multirow}
\usepackage[table]{xcolor}
\usepackage{algorithm}
\usepackage{algorithmic}
\usepackage{etoc}

\usepackage{booktabs}
\usepackage{graphicx}
\usepackage{multirow}
\usepackage{amssymb}
\usepackage{pifont}
\usepackage{array}
\usepackage{makecell}

\usepackage{booktabs}
\usepackage{multirow}
\usepackage{amssymb}
\usepackage{pifont}
\usepackage{array}
\usepackage{graphicx}
\usepackage{makecell}

\newcommand{\cmark}{\textbf{\ding{51}}}
\newcommand{\xmark}{\textbf{\ding{55}}}

\definecolor{t2e_red}{RGB}{239,99,75}
\definecolor{t2e_blue}{RGB}{99,113,250}
\definecolor{t2e_green}{RGB}{0,180,139}
\definecolor{t2e_gray}{RGB}{165,165,165}
\definecolor{redlink}{RGB}{239,99,75}
\definecolor{t2e_purple}{RGB}{155,89,182}

\usepackage[capitalize,noabbrev]{cleveref}

\theoremstyle{plain}

\theoremstyle{definition}

\theoremstyle{remark}

\usepackage[textsize=tiny]{todonotes}

\icmltitlerunning{Q-Hawkeye: Reliable Visual Policy Optimization for Image Quality Assessment}

\begin{document}
\twocolumn[
  \icmltitle{Q-Hawkeye: Reliable Visual Policy Optimization for Image Quality Assessment
}



\begin{icmlauthorlist}
  \icmlauthor{Wulin Xie$^*$}{casia,ucas,amap}
  \icmlauthor{Rui Dai$^\dagger$}{amap}
  \icmlauthor{Ruidong Ding}{amap}
  \icmlauthor{Kaikui Liu}{amap}
  \icmlauthor{Xiangxiang Chu}{amap}
  \icmlauthor{Xinwen Hou}{casia}
  \icmlauthor{Jie Wen}{hit}
\end{icmlauthorlist}
\icmlaffiliation{casia}{Institute of Automation, Chinese Academy of Sciences, Beijing, China}
\icmlaffiliation{ucas}{University of Chinese Academy of Sciences, Beijing, China}
\icmlaffiliation{amap}{Amap, Alibaba Group, Hangzhou, China}
\icmlaffiliation{hit}{Harbin Institute of Technology, Shenzhen, China}
\icmlcorrespondingauthor{Xinwen Hou}{xinwen.hou@ia.ac.cn}
\icmlcorrespondingauthor{Jie Wen}{jiewen@126.com}
\icmlkeywords{Machine Learning, ICML}
\vskip 0.3in
]
\printAffiliationsAndNotice{$^*$Work done during an internship at Amap. $^\dagger$Project Leader.}

\begin{abstract}
Image Quality Assessment (IQA) predicts perceptual quality scores consistent with human judgments. Recent RL-based IQA methods built on MLLMs focus on generating visual quality descriptions and scores, ignoring two key reliability limitations: (i) although the model’s prediction stability varies significantly across training samples, existing GRPO-based methods apply uniform advantage weighting, thereby amplifying noisy signals from unstable samples in gradient updates; (ii) most works emphasize text-grounded reasoning over images while overlooking the model’s visual perception ability of image content. In this paper, we propose Q-Hawkeye, an RL-based reliable visual policy optimization framework that redesigns the learning signal through unified Uncertainty-Aware Dynamic Optimization and Perception-Aware Optimization. Q-Hawkeye estimates predictive uncertainty using the variance of predicted scores across multiple rollouts and leverages this uncertainty to reweight each sample’s update strength, stabilizing policy optimization. To strengthen perceptual reliability, we construct paired inputs of degraded images and their original images and introduce an Implicit Perception Loss that constrains the model to ground its quality judgments in genuine visual evidence. Extensive experiments demonstrate that Q-Hawkeye outperforms state-of-the-art methods and generalizes better across multiple datasets. Our dataset and code are available at \url{https://github.com/AMAP-ML/Q-Hawkeye}.
\end{abstract}    

\section{Introduction}
\label{sec:intro}
Image Quality Assessment (IQA) aims to predict perceptual quality scores consistent with human subjective judgments and serves as a fundamental component in vision applications such as enhancement and AIGC quality control~\cite{SPAQ, wu2024openendedvisualqualitycomparison, FingER2,VMBench}. Depending on whether a reference is available, IQA is typically categorized into full-reference (FR) and no-reference (NR) settings~\cite{1284395, LD-RPS}, with NR-IQA being especially practical in real-world scenarios~\cite{RealQA}. With the rapid progress of multimodal large language models (MLLMs)~\cite{diao2025temporal, MM-RLHF, diao2026addressing}, recent IQA methods have leveraged their ability to jointly model low-level degradations and high-level perceptual attributes, often producing text-grounded rationales alongside scalar scores~\cite{wu2024comprehensivestudymultimodallarge}. Recently, reinforcement learning has been explored as a post-training paradigm to further improve reasoning and generalization. For instance, Q-Insight~\cite{Q-insight} applies GRPO~\cite{DeepSeekMath} with verifiable rewards to jointly optimize score prediction and distortion awareness, while VisualQuality-R1~\cite{VisualQuality-R1} frames NR-IQA as a pairwise ranking problem, performs multiple rollouts per image pair, and uses a Thurstone-style probabilistic formulation with continuous fidelity rewards to induce reasoning-aware quality ranking behavior.

\begin{figure}
  \centering
\includegraphics[width=1.0\linewidth]{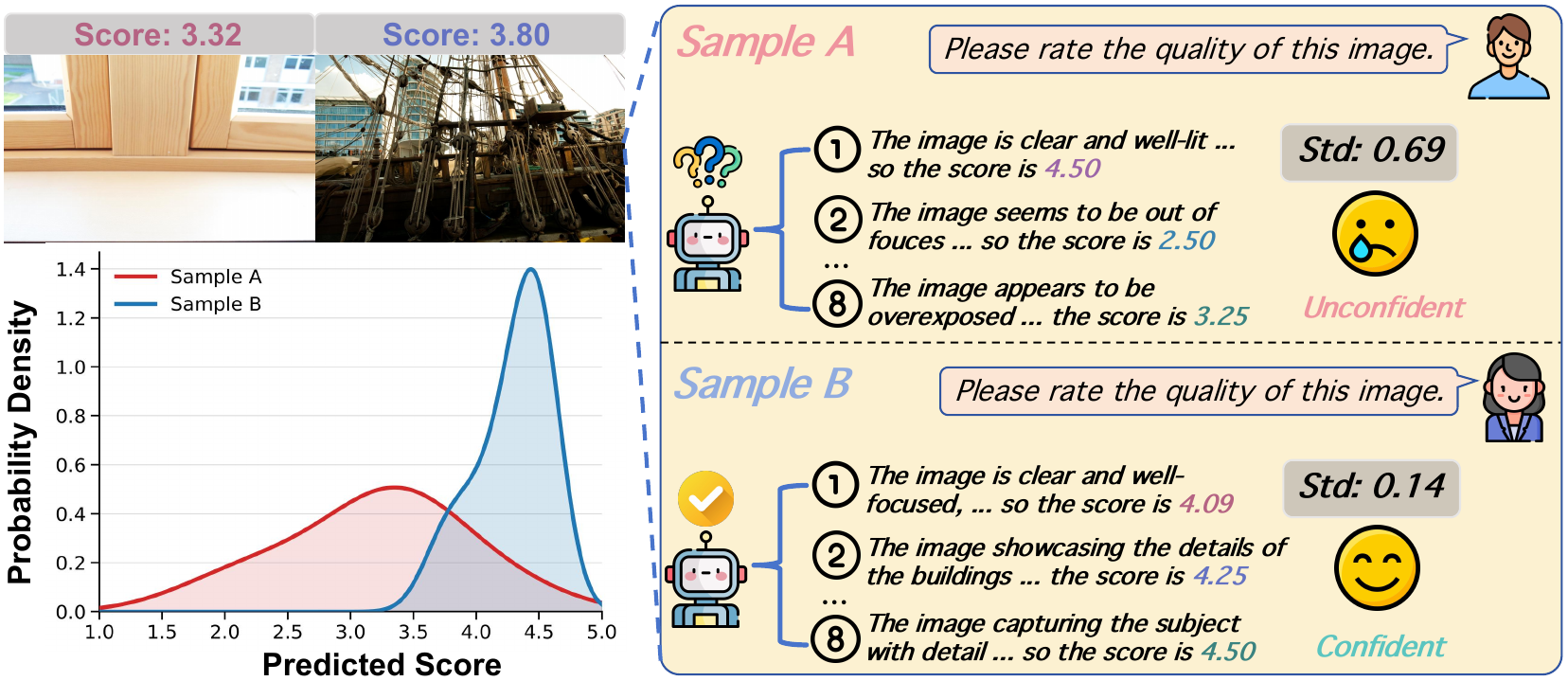}
\caption{Illustration of predictive uncertainty during training. Sample A shows a high-variance score distribution with inconsistent reasoning, while Sample B shows a low-variance distribution with consistent reasoning.}  \label{fig:uncertainty_example}
\vspace{-15pt}  
\end{figure}

\begin{figure*}[t]
  \centering
\includegraphics[width=1.0\linewidth]{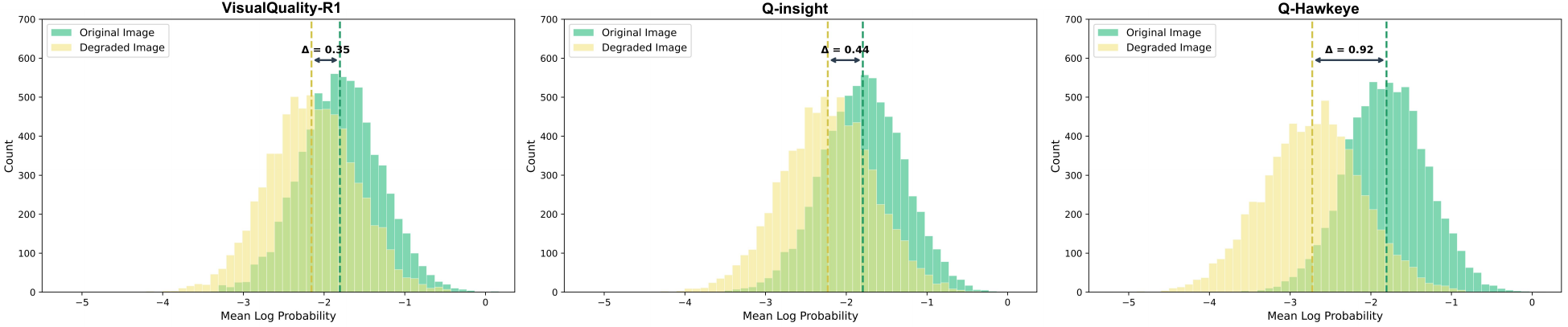}
\caption{\textbf{Visual perception analysis of existing state-of-the art methods and our Q-Hawkeye on the KonIQ dataset.} For each method, we sample responses conditioned on the original images, and compute the mean log-probability of the same responses under the \textcolor[HTML]{29BD7D}{original} and \textcolor[HTML]{F2E582}{degraded} conditions. Histograms show the distributions and the mean gap between them.}
 \label{fig:distribution_1}
\vspace{-3pt}  
\end{figure*}

Despite this progress, we observe two persistent limitations in RL-based IQA methods. Firstly, rollout score distributions vary markedly across images; some samples generate consistent predictions, while others predict scores with broad and unstable distributions (Fig.~\ref{fig:uncertainty_example}). This difference reflects predictive uncertainty: for certain images, the model has not formed a stable quality judgment. However, existing GRPO-based methods~\cite{Q-insight, Q-Ponder, VisualQuality-R1} commonly apply uniform update strength across samples, causing unstable instances to inject noise into gradient updates and undermining optimization reliability~\cite{wang2023selfconsistencyimproveschainthought, kuhn2023semanticuncertaintylinguisticinvariances, SEED-GRPO}. Secondly, existing methods focus on the model's textual reasoning, score regression, or ranking ability, but ignore the model’s visual perception ability to image content and degradation, which is crucial for visual quality assessment~\cite{PAPO}. As shown in Fig.~\ref{fig:distribution_1} and Fig.~\ref{fig:distribution_2}, images with degradation still receive scores and outputs close to relatively clean images, suggesting that in the absence of explicit constraints, the model’s scores and outputs may not be fully grounded in the visual content. In other words, the model can partially rely on dataset regularities or language priors rather than image-intrinsic evidence, restricting the model's reliability and generalization.

To address these challenges, we propose Q-Hawkeye, a new RL training paradigm for the IQA task that redesigns the learning signal from the perspectives of model uncertainty and visual perceptual ability. Firstly, we introduce an Uncertainty-Aware Dynamic Optimization strategy that performs multiple rollouts per image and estimates uncertainty from the variance of the predicted scores. This uncertainty is then used to adaptively reweight sample-wise update strength, upweighting low-uncertainty samples to consolidate reliable judgments while downweighting high-uncertainty ones to avoid aggressive updates where the policy remains unstable. Secondly, we introduce a Perception-Aware Optimization strategy. Specifically, we first construct paired inputs of original images and their degraded counterparts across diverse distortion types, and introduce an Implicit Perception Loss over the policy outputs under the two visual conditions. This loss encourages the model to produce distinguishable output distributions for original and degraded inputs, thereby strengthening the model’s perceptual ability to visual contents.

In summary, our contributions are:
\begin{itemize}
    \item We propose Q-Hawkeye, a reliable visual policy optimization framework for IQA, which introduces unified Uncertainty-Aware Dynamic Optimization and Perception-Aware Optimization strategies that jointly improve the reliability of image quality assessment.

    \item Unlike prior methods that focus on textual reasoning and description abilities, we explicitly explore the visual perceptual capability of MLLMs for IQA. Specifically, we first construct original-degraded image pairs and then introduce an Implicit Perception Loss that drives visually grounded and separable quality judgments, enhancing the model's perceptual ability to image's visual content.

    \item Extensive experiments demonstrate our Q-Hawkeye outperforms existing state-of-the-art IQA methods with improved robustness and generalization across datasets and degradation conditions.
\end{itemize}



\section{Related Work}
\label{sec:related_work}

\textbf{Score-based IQA Methods} predict continuous quality scores for images. Approaches fall into traditional non-MLLM regressors and recent MLLM-based predictors. Non-MLLM methods cover both full-reference and no-reference IQA, evolving from hand-crafted statistical features to deep CNN or Transformer regressors such as~\cite{NIMA, NIQE, CLIP-IQA, ManIQA, BRISQUE, HyperIQA, MUSIQ}. MLLM-based methods instead leverage large vision--language models for scoring, including discrete quality level prediction~\cite{Q-Align, RealQA, Q-Scorer, Q-Mamba}, distribution-based supervision~\cite{DeQA-Score}, reinforcement-driven quality and degradation modeling~\cite{Q-insight}, and pairwise ranking with preference optimization~\cite{VisualQuality-R1}, aiming to align scores with human judgments and generalize across diverse distortions.

\textbf{Ranking-based IQA Methods} view perceptual quality as a relative preference. Instead of directly regressing a score, the model learns which image in a pair is of higher quality. Early work such as~\cite{RankIQA, dipIQ} demonstrated that quality-discriminable image pairs can be generated automatically and used to train learning-to-rank models. Subsequent approaches extended this idea with multitask or uncertainty-aware objectives that jointly model quality, content, and distortion type, and with fidelity-style pairwise losses that help align synthetic and in-the-wild degradations, improving robustness across datasets~\cite{RankNet, Zhang_2021}. Recently, multimodal ranking-based methods such as Compare2Score~\cite{Compare2Score} train a vision--language comparator to express relative quality preferences and then convert these pairwise preferences into continuous scores via probabilistic inference against anchor images, effectively bridging preference reasoning and single-image scoring.

\textbf{Reinforcement learning (RL)} has recently become a core paradigm for training reasoning-oriented LLMs~\cite{diao2025soundmind, HS-STaR, Tree-GRPO}. DeepSeek-R1~\cite{DeepSeek-R1} introduces Group Relative Policy Optimization (GRPO), which eliminates value models by using group-relative baselines from multiple rollouts, enabling scalable training with rule-based rewards. Building on GRPO, variants such as GPG~\cite{GPG} directly optimizes the original RL objective while improving advantage estimation for more efficient training than GRPO. DAPO~\cite{DAPO} dynamically filter consistently correct or consistently incorrect trajectories to focus updates on informative samples, while EMPO~\cite{EMPO} incorporates semantic entropy as an uncertainty signal to guide optimization, highlighting the role of uncertainty-aware policy learning. In multimodal settings, RL has been extended to visual grounding and multimodal reasoning~\cite{Visual-RFT, R1-VL, PAPO}. More recent work further exploits RL signals to enhance perceptual grounding by either aligning model outputs with fine-grained visual quality under verifiable rewards~\cite{Visionary-R1}, or encouraging perception-enhancing behaviors, thereby strengthening the model’s intrinsic visual perception capabilities~\cite{Pixel-Reasoner}.
   
\begin{figure*}[h]
  \centering
  \vspace{-3pt}  
\includegraphics[width=1.0\linewidth]{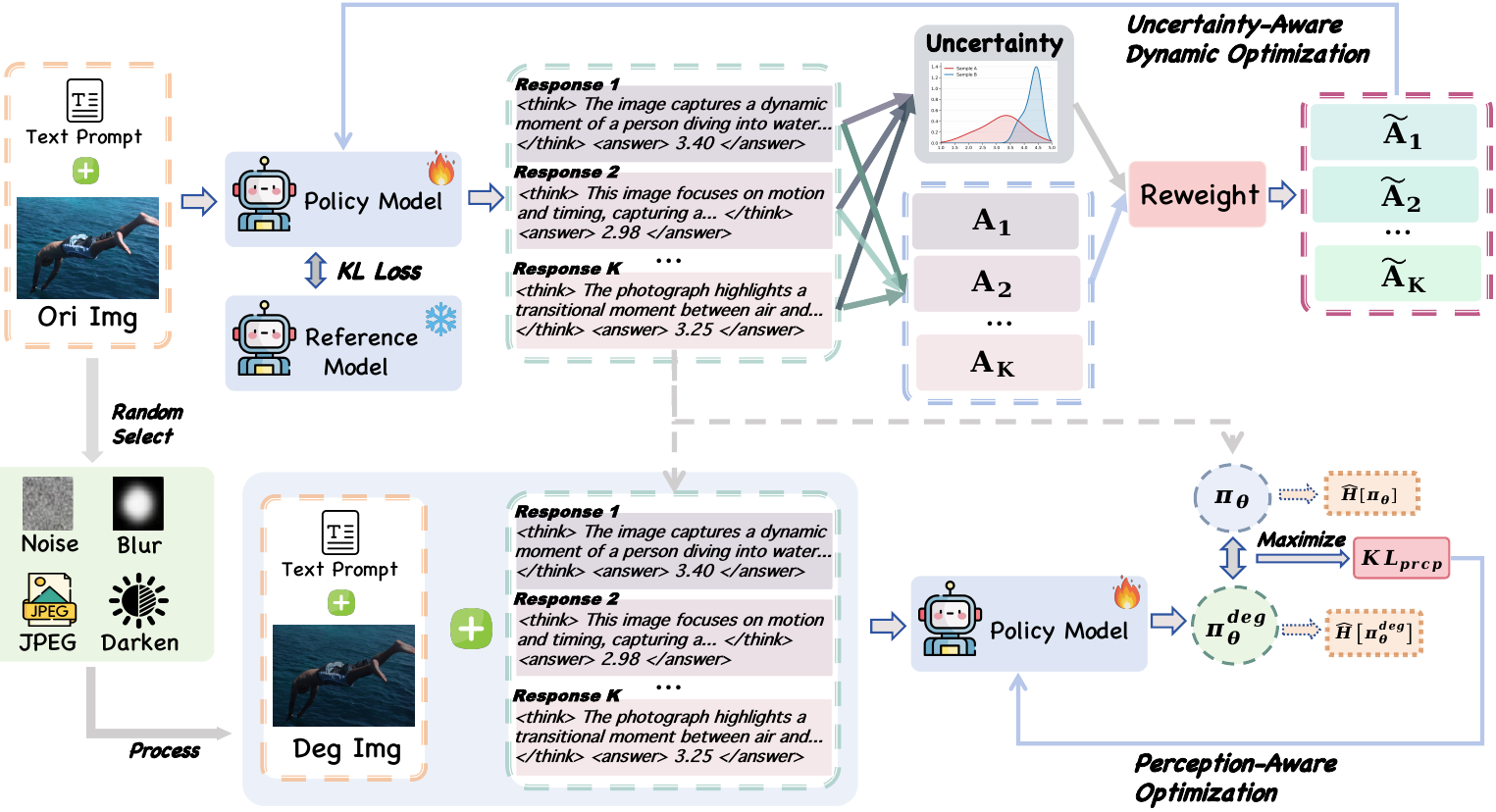}
\caption{Overview of the proposed Q-Hawkeye framework. For each image--prompt pair, the policy model produces $K$ reasoning trajectories and quality scores. The \textbf{Uncertainty-Aware Dynamic Optimization} module computes the score variance as an uncertainty signal and uses it to rescale sample-wise advantages. In parallel, \textbf{Perception-Aware Optimization} constructs original--degraded image pairs and maximizes the KL divergence between their output distributions, enforcing sensitivity to visual degradations.}
\label{fig:visualization}
\vspace{-10pt}  
\end{figure*}

\section{Method}
\subsection{Preliminaries}

Image Quality Assessment (IQA) aims to learn a mapping from images to perceptual quality such that model predictions align with human subjective judgments.  
Given a training dataset $\mathcal{D} = \{(I, q, y)\}$, each triplet consists of an input image $I$, a text prompt $q$, and a Mean Opinion Score (MOS) $y$.  
We adopt a pretrained vision-language model as the policy $\pi_\theta(\cdot \mid I, q)$, which generates a textual output $o$ containing both the reasoning process and the final answer conditioned on $(I, q)$.  
By parsing the \texttt{<answer>} field in $o$, we obtain a scalar quality prediction $\hat{y} \in [1, 5]$.

Since ground-truth MOS are continuous and inherently subjective, using a sparse binary ``correct/incorrect'' reward would cause most rollouts to receive near-zero feedback in early training, providing little signal for improving suboptimal predictions.  
To obtain smoother and denser supervision, following~\cite{Q-Ponder}, we define a score-based continuous reward that depends on the deviation between the predicted and MOS:
\begin{equation}
    R_{\mathrm{acc}}
    = \exp\left(-\frac{\lvert \hat{y} - y \rvert}{\alpha}\right),
    \quad
    R_{\mathrm{acc}} \in (0, 1),
\end{equation}
where $\alpha$ is a tolerance hyperparameter, smaller $\alpha$ imposes stronger penalties on prediction errors.

To ensure that the reasoning trace and final answer remain parseable and to stabilize training, we additionally introduce a format reward that enforces a structured output pattern with \texttt{<think>...</think>} and \texttt{<answer>...</answer>} blocks:
\begin{equation}
    R_{\mathrm{fmt}} =
    \begin{cases}
        1, & \text{if the output format satisfies the requirements},\\
        0, & \text{otherwise}.
    \end{cases}
\end{equation}
The total reward for each rollout is then defined as a sum of the two components:
\begin{equation}
    r = R_{\mathrm{total}}
      = R_{\mathrm{acc}}
      +  R_{\mathrm{fmt}},
\end{equation}

\noindent\textbf{Group Relative Policy Optimization.}
We adopt Group Relative Policy Optimization (GRPO) as our post-training algorithm.  
For any sample $(I, q)$, we first draw a group of $K$ rollouts from the old policy
$\pi_{\theta_{\mathrm{old}}}$:
\begin{equation}
    \{o_k\}_{k=1}^{K} \sim \pi_{\theta_{\mathrm{old}}}(\cdot \mid I, q),
\end{equation}
and compute the corresponding rewards $\{r_k\}_{k=1}^{K}$.  
Instead of relying on a separate value model, GRPO constructs relative advantages directly from group-wise rewards to encourage the policy to favor higher-reward outputs for the same input.  
Concretely, we normalize the rewards within the group to obtain advantages as follows:
\begin{equation}
    A_k
    = \frac{r_k - \mu(\mathbf{r})}{\sigma(\mathbf{r})},
    \quad
    \mathbf{r} = [r_1, \dots, r_K],
\end{equation}
where $\mu(\cdot)$ and $\sigma(\cdot)$ denote the mean and standard deviation over the group.

We then define the importance sampling ratio
\begin{equation}
    \rho_k(\theta)
    = \frac{\pi_\theta(o_k \mid I, q)}{\pi_{\theta_{\mathrm{old}}}(o_k \mid I, q)},
\end{equation}
and adopt a clipped surrogate objective to control the update magnitude.  
In addition, a KL regularization is used to limit distributional drift.  
The overall GRPO optimization objective is written as:
\begin{equation}
{\small
\begin{aligned}
\mathcal{L}_{\mathrm{GRPO}}(\theta)
&= -\frac{1}{K}\sum_{k=1}^{K}\min\!\Big(
\rho_k(\theta)A_k,\;
\mathrm{clip}\!\big(\rho_k(\theta),\,1\pm\varepsilon\big) A_k\Big) \\
&\quad + \beta\, D_{\mathrm{KL}}\!\Big(
\pi_\theta(\cdot \mid I,q)\;\big\|\;\pi_{\mathrm{ref}}(\cdot \mid I,q)
\Big).
\end{aligned}
}
\end{equation}
Specifically:
\begin{equation}
\begin{aligned}
&D_{\mathrm{KL}}\!\Big(
\pi_\theta(\cdot \mid I,q)\,\big\|\,\pi_{\mathrm{ref}}(\cdot \mid I,q)
\Big) \\
&=\mathbb{E}_{o \sim \pi_\theta(\cdot \mid I,q)}
\Big[
\log \pi_\theta(o \mid I,q)
-
\log \pi_{\mathrm{ref}}(o \mid I,q)
\Big].
\end{aligned}
\end{equation}
where $\varepsilon$ is the clipping threshold and $\beta$ controls the strength of KL regularization.

\subsection{Uncertainty-Aware Dynamic Optimization}

\paragraph{Motivation.}
During the training stage, we observe that, under the same $K$ rollouts, different images induce markedly different dispersions in the predicted quality scores (Fig.~\ref{fig:uncertainty_example}): for some samples, the scores concentrate within a narrow range, whereas for others they fluctuate more broadly. Motivated by prior works showing that disagreement among multiple sampled outputs can be interpreted as predictive uncertainty~\cite{SEED-GRPO, wang2023selfconsistencyimproveschainthought, kuhn2023semanticuncertaintylinguisticinvariances}, we regard the dispersion of per-rollout predicted scores as an uncertainty estimate for the model's quality prediction on a given image.

To prevent high-uncertainty samples from dominating optimization, we downweight their policy-gradient contributions. For a single rollout, a GRPO-style objective yields:
\begin{equation}
    L(\theta) \propto \rho(\theta) A,
\end{equation}
and the corresponding policy gradient is
\begin{equation}
    \nabla_{\theta} L(\theta)
    \propto
    \rho(\theta)\, A\, \nabla_{\theta}\log \pi_{\theta}(o \mid I, q).
\end{equation}
Thus, the advantage $A$ linearly scales the gradient magnitude. As a result, we introduce an uncertainty-aware weight $w(u)\in(0,1]$ and rescale the advantage as
\begin{equation}
    \tilde{A} = w(u)\,A,
\end{equation}
which can be interpreted as a sample-specific update strength: low-uncertainty samples receive stronger reinforcement, while high-uncertainty ones are suppressed to reduce gradient variance and noisy amplification.

\paragraph{Uncertainty-aware advantage reweighting.}
For each training sample $(I,q)$, we draw $K$ rollouts from the old policy:
\begin{equation}
    \{o_k\}_{k=1}^{K} \sim \pi_{\theta_{\mathrm{old}}}(\cdot \mid I, q),
\end{equation}
and parse a predicted quality score from each rollout, yielding $\{\hat{y}_k\}_{k=1}^{K}$.
We define the sample-level uncertainty as the within-group variance:
\begin{equation}
    \bar{y}=\frac{1}{K}\sum_{k=1}^{K}\hat{y}_k,
    \qquad
    u=\frac{1}{K}\sum_{k=1}^{K}(\hat{y}_k-\bar{y})^2.
\end{equation}
Since scores are linearly mapped to $[y_{\min},y_{\max}]$ (e.g., $[1,5]$), we normalize $u$ by the maximal variance bound $(\Delta_y^2/4)$ with $\Delta_y=y_{\max}-y_{\min}$:
\begin{equation}
    \tilde{u}
    =
    \mathrm{min}\!\left(
        \frac{u}{\Delta_y^2/4 + \epsilon_u},
        1
    \right),
\end{equation}
where $\epsilon_u>0$ is a small constant for numerical stability, and the min operation ensures that the normalized score remains within the range $[0,1]$, preventing out-of-range predicted scores from distorting the uncertainty estimation.

We then map $\tilde{u}$ to a downweighting factor
\begin{equation}
    w=\exp(-\tau \tilde{u}),
\end{equation}
with temperature $\tau$ controlling the suppression strength.

In GRPO, rollout-wise rewards $\{r_k\}_{k=1}^{K}$ are used to compute group-relative advantages $\{A_k\}_{k=1}^{K}$. We apply the same sample-wise weight $w$ to all rollouts in the group:
\begin{equation}
    \tilde{A}_k = w \cdot A_k.
\end{equation}
This preserves strong updates for stable samples while attenuating ambiguous ones, improving training stability and reducing the impact of noisy signals.

\subsection{Perception-Aware Optimization}


\paragraph{Degradation Data Construction.} To encourage the model to rely more on visual evidence in quality assessment and to improve its sensitivity to degradation changes, we construct paired original--degraded data for training. Concretely, for each image $I$ in the training set, we randomly sample one degradation type from the set $\{\emph{Noise}, \emph{Blur}, \emph{JPEG}, \emph{Darken}\}$ and apply it with a fixed parameter to obtain a degraded version $I^{\mathrm{deg}}$. Each training sample is thus extended to a pair of visual inputs $(I, I^{\mathrm{deg}})$ while sharing the same task prompt $q$. Detailed instructions and examples are provided in the Appendix~\ref{sec:app_degraded} and~\ref{sec:app_vis}.

However, a key issue is that some degradations may be too weak to induce a noticeable perceptual change. For example, applying additional blur to an already blurred image, in which case the original and degraded images fail to form an effective contrastive signal. To avoid this, we introduce a double-check filter strategy during data construction. Specifically, we first employ a strong MLLM (e.g., GPT-4o) to judge, for each pair $(I, I^{\mathrm{deg}})$, whether there exists a discernible visual difference. For pairs that the MLLM cannot reliably distinguish, we further judge them by human experts. If human experts also fail to perceive a discernible quality difference, we regard the degraded sample as not meeting the ``effective contrast'' requirement and resample the degradation type to regenerate $I^{\mathrm{deg}}$. We repeat this process until all pairs in the training set can be consistently distinguished by both the MLLM and human experts, constructing the final origin--degraded training set.

\paragraph{Implicit Perception Loss.}
Given the paired data $(I, I^{\mathrm{deg}})$, our goal is to encourage the model to evaluate visual quality based on visual evidence. If the model truly perceives the image content during assessment, then changing the input from the original image $I$ to the degraded image $I^{\mathrm{deg}}$ should lead to a noticeable change in the reasoning and predicted score. Conversely, if the model is unable to distinguish the two visual conditions, then with the same prompt, the two inputs are effectively equivalent to the model, so the output distribution should change minimally. We therefore introduce an \emph{Implicit Perception Loss} that measures the discrepancy between the policy distributions under the two visual conditions and forces the model to remain sensitive to visual degradations.

For each sample $(I, q)$, the GRPO rollout process produces a group of outputs
$\{o_k\}_{k=1}^{K}$, where each $o_k$ is a full sequence containing both reasoning and the final score. For every $o_k$, we compute its log-probability under the original and degraded conditions:
\begin{equation}
    \log \pi_\theta(o_k \mid I, q),
    \qquad
    \log \pi_\theta(o_k \mid I^{\mathrm{deg}}, q).
\end{equation}
We define the perception ratio
\begin{equation}
    p_{\mathrm{ipl}}^{(k)}(\theta)
    =
    \frac{\pi_\theta(o_k \mid I, q)}
         {\pi_\theta(o_k \mid I^{\mathrm{deg}}, q)}.
\end{equation}
When the model's judgment strongly depends on visual information, degradation should significantly change the output distribution, causing this ratio to deviate from $1$~\cite{PAPO}. We further quantify the discrepancy between the two visual conditions using the KL divergence and treat it as an implicit perception signal:
\begin{equation}
\label{eq:kl_prcp}
\begin{aligned}
    &D_{\mathrm{KL}}\!\Big(
        \pi_\theta(\cdot \mid I, q)
        \,\Big\|\,
        \pi_\theta(\cdot \mid I^{\mathrm{deg}}, q)
    \Big)
    \\
    &=
    \mathbb{E}_{o \sim \pi_\theta(\cdot \mid I, q)}
    \Big[
        \log \pi_\theta(o \mid I, q)
        - \log \pi_\theta(o \mid I^{\mathrm{deg}}, q)
    \Big].
\end{aligned}
\end{equation}
We maximize the KL discrepancy in Eq~\eqref{eq:kl_prcp} to encourage the policy to react to degradations, so that the model can generate distinguishable output distributions under the two visual conditions $I$ and $I^{\mathrm{deg}}$.

However, the model may artificially increase the KL term by making its outputs more random or diffuse, rather than by learning visually grounded and consistent judgments, which can degrade performance or even destabilize training~\cite{PAPO}. To suppress such degenerate behaviors, we introduce a \emph{double entropy regularization} that constrains the entropy of the policy under both pristine and degraded conditions, encouraging sufficiently sharp and stable output distributions. Using the rollout log-probabilities to estimate entropy, we define:
\begin{equation}
\label{eq:double_entropy}
    \mathcal{L}_{\mathrm{ent}}(\theta)
    =
    \eta_1\, \hat{H}\!\big(\pi_\theta(\cdot \mid I, q)\big)
    +
    \eta_2\, \hat{H}\!\big(\pi_\theta(\cdot \mid I^{\mathrm{deg}}, q)\big),
\end{equation}
where $\eta_1, \eta_2$ are weighting coefficients and $\hat{H}(\cdot)$ denotes an estimate of the policy entropy based on the $K$ rollouts. In practice, we reuse the same rollout group $\{o_k\}\sim\pi_{\theta_{\mathrm{old}}}(\cdot\mid I,q)$ and obtain $\log\pi_\theta(o_k\mid I^{\mathrm{deg}},q)$ via a paired forward pass to estimate $\hat H(\cdot)$ and $D_{\mathrm{KL}}$; the resulting $\hat H$ under $I^{\mathrm{deg}}$ is an off-policy cross-entropy proxy, which provides a stable regularization signal against high-entropy collapse. Specifically, given rollouts $\{o_k\}_{k=1}^K$ we approximate the entropy under a visual condition $x \in \{I, I^{\mathrm{deg}}\}$ as:
\begin{equation}
\hat{H}\!\big(\pi_\theta(\cdot \mid x, q)\big)
= - \frac{1}{K} \sum_{k=1}^K \log \pi_\theta(o_k \mid x, q).
\end{equation}

\paragraph{Overall Optimization Objective.}
Finally, the overall optimization objective is:
\begin{equation}
\label{eq:total_objective}
\small 
\begin{aligned}
\mathcal{L}_{\mathrm{Total}}(\theta)
&= \mathbb{E}_{\substack{(I, q, I^{\mathrm{deg}}) \sim \mathcal{D} \\ \{o_k\}_{k=1}^{K} \sim \pi_{\theta_{\mathrm{old}}}(\cdot \mid I, q)}}
\\
&\Bigg[
    -\frac{1}{K}\sum_{k=1}^{K}
    \min\Bigl(
        \rho_k(\theta)\,\tilde{A}_k,\,
        \operatorname{clip}\bigl(\rho_k(\theta), 1\pm\varepsilon\bigr)\,\tilde{A}_k
    \Bigr)
\\
&\qquad 
    + \beta\, D_{\mathrm{KL}}\Bigl(
        \pi_\theta(\cdot \mid I, q)
        \,\Big\|\,
        \pi_{\mathrm{ref}}(\cdot \mid I, q)
    \Bigr)
\\
&\qquad 
    - \gamma\, D_{\mathrm{KL}}\Bigl(
        \pi_\theta(\cdot \mid I, q)
        \,\Big\|\,
        \pi_\theta(\cdot \mid I^{\mathrm{deg}}, q)
    \Bigr)
\\
&\qquad 
    + \eta_1\, \hat{H}\bigl(\pi_\theta(\cdot \mid I, q)\bigr)
    + \eta_2\, \hat{H}\bigl(\pi_\theta(\cdot \mid I^{\mathrm{deg}}, q)\bigr)
\Bigg].
\end{aligned}
\end{equation}
where $\tilde{A}_k$ are the uncertainty-reweighted advantages, $\varepsilon$ is the clipping threshold, $\beta$ controls the KL regularization with respect to the reference policy $\pi_{\mathrm{ref}}$, $\gamma$ controls the strength of the implicit perception term, and $\eta_1, \eta_2$ control the double entropy regularization. The overall algorithmic workflow can be found in Algorithm~\ref{alg:qhawkeye}.

\section{Experiments}
\subsection{Experimental Setup}

\textbf{Datasets and Metrics.}
Our Q-Hawkeye focuses on the score regression task, where the model predicts a continuous Mean Opinion Score (MOS) for an image, and we evaluate on various IQA datasets. Following~\cite{DeQA-Score}, we split KonIQ into training and test sets with approximately 7k training images, and training is performed only on KonIQ~\cite{KonIQ}. All MOS labels are linearly mapped to the range $[1, 5]$ for supervision. To assess generalization, we evaluate on both the test set of KonIQ (in-distribution) and a set of out-of-distribution IQA datasets spanning four categories:
(a) in-the-wild images, including SPAQ~\cite{SPAQ}, LIVE-Wild~\cite{LiveW}, and FLIVE~\cite{FLIVE};
(b) synthetic distortion datasets, including KADID-10K~\cite{KIDID-10K} and CSIQ~\cite{CSIQ};
(c) model-processed distortions, represented by PIPAL~\cite{gu2020pipallargescaleimagequality};
and (d) AI-generated images from AGIQA-3K~\cite{li2023agiqa3kopendatabaseaigenerated}.
Following prior work~\cite{MUSIQ, Q-Align},
We report the Pearson Linear Correlation Coefficient (PLCC) and the Spearman Rank-Order Correlation Coefficient (SRCC) between predicted scores and human MOS.

\textbf{Baselines.}
We compare against three groups of IQA approaches.
(1) Handcrafted no-reference quality models, including NIQE~\cite{NIQE} and BRISQUE~\cite{BRISQUE}.
(2) Deep learning-based IQA models, including NIMA~\cite{NIMA}, MUSIQ~\cite{MUSIQ}, CLIP-IQA+~\cite{CLIP-IQA}, and ManIQA~\cite{ManIQA}.
(3) Recent MLLM-based methods that leverage large vision-language models for perceptual quality assessment, including
Compare2Score~\cite{Compare2Score},
Qwen-SFT~\cite{Qwen2.5-VL},
Q-Align~\cite{Q-Align},
DeQA~\cite{DeQA-Score},
Q-Insight~\cite{Q-insight},
and VisualQuality-R1~\cite{VisualQuality-R1}.

\textbf{Implementation Details.}
Our model is optimized based on Qwen2.5-VL-7B~\cite{Qwen2.5-VL}, where we set $K = 8$ rollouts per input. We set $\alpha = 0.30$, KL regularization weight $\beta = 1 \times 10^{-3}$, $\tau = 0.2$, $\gamma = 5 \times 10^{-4}$, and $\eta_1 = \eta_2 = 1 \times 10^{-4}$.
We use AdamW~\cite{AdamW} with an initial learning rate of $5 \times 10^{-6}$, a total batch size of 32, and train for 15 epochs. All experiments are conducted on 8 NVIDIA H20 GPUs.

\begin{table*}[t]
\centering
\caption{\textbf{PLCC / SRCC comparison of our Q-Hawkeye with existing IQA methods on eight IQA benchmarks.} Methods are grouped as \textcolor{t2e_blue}{$\bullet$~\textbf{Handcrafted}}, \textcolor{t2e_green}{$\bullet$~\textbf{Non-MLLM Deep-Learning}}, and \textcolor{t2e_red}{$\bullet$~\textbf{MLLM-Based Methods}}. The best and second-best results for each dataset are marked with \textbf{bold} and \underline{underline}, respectively, and \textit{Avg} denotes the average performance over all datasets.}
\vspace{0.1cm}
\resizebox{\linewidth}{!}{
\begin{tabular}{r|r|cccccccc|c}
\toprule
\textbf{Method} & \textbf{Venue} & \textbf{KonIQ} & \textbf{SPAQ} & \textbf{KADID} & \textbf{PIPAL} & \textbf{LIVE-Wild} & \textbf{AGIQA-3K} & \textbf{CSIQ} & \textbf{FLIVE} & \textbf{Avg} \\
\midrule\midrule
\rowcolor{t2e_blue!15}\multicolumn{11}{l}{\textcolor{t2e_blue}{$\bullet$~\textbf{Handcrafted}}}\\
NIQE & \textcolor{gray}{{SPL'12}} 
& $53.3/53.0$ & $67.9/66.4$ & $46.8/40.5$ & $19.5/16.1$ & $49.3/44.9$ & $56.0/53.3$ & $71.8/62.8$ & $14.7/10.0$ & $46.2/43.4$ \\
BRISQUE & \textcolor{gray}{{TIP'12}} 
& $22.5/22.6$ & $49.0/40.6$ & $42.9/35.6$ & $26.7/23.2$ & $36.1/31.3$ & $54.1/49.7$ & $74.0/55.6$ & $10.8/5.4$ & $39.5/33.5$ \\
\midrule
\rowcolor{t2e_green!13}\multicolumn{11}{l}{\textcolor{t2e_green}{$\bullet$~\textbf{Non-MLLM Deep-Learning}}}\\
NIMA & \textcolor{gray}{{TIP'18}} 
& $89.6/85.9$ & $83.8/85.6$ & $53.2/53.5$ & $39.0/39.9$ & $81.4/77.1$ & $71.5/65.4$ & $69.5/64.9$ & $56.1/46.7$ & $68.0/64.9$ \\
HyperIQA & \textcolor{gray}{{CVPR'20}} 
& $91.7/90.6$ & $79.1/78.8$ & $50.6/46.8$ & $41.0/40.3$ & $77.2/74.9$ & $70.2/64.0$ & $75.2/71.7$ & $48.5/38.3$ & $66.7/63.2$ \\
DBCNN & \textcolor{gray}{{ICSIPA'19}} 
& $88.4/87.5$ & $81.2/80.6$ & $49.7/48.4$ & $38.4/37.1$ & $77.3/73.5$ & $73.0/70.1$ & $58.6/57.2$ & $48.5/38.5$ & $63.1/61.6$ \\
MUSIQ & \textcolor{gray}{{ICCV'21}} 
& $92.4/92.9$ & $86.8/86.3$ & $57.5/55.6$ & $43.1/43.1$ & $78.9/83.0$ & $72.2/63.0$ & $71.1/71.0$ & $56.5/46.7$ & $69.8/67.0$ \\
CLIP-IQA+ & \textcolor{gray}{{AAAI'23}} 
& $90.9/90.5$ & $86.6/86.4$ & $65.3/65.4$ & $43.7/43.1$ & $80.5/75.2$ & $73.9/73.7$ & $75.2/73.7$ & $51.2/43.1$ & $71.8/68.9$ \\
ManIQA & \textcolor{gray}{{CVPR'22}} 
& $84.9/83.4$ & $76.8/75.8$ & $49.9/46.5$ & $45.7/45.2$ & $84.9/83.2$ & $72.3/66.3$ & $62.3/62.7$ & $51.2/40.1$ & $65.9/64.2$ \\
\midrule
\rowcolor{t2e_red!15}\multicolumn{11}{l}{\textcolor{t2e_red}{$\bullet$~\textbf{MLLM-Based Methods}}}\\
Compare2Score & \textcolor{gray}{{NIPS'24}} 
& $92.3/91.0$ & $86.7/86.0$ & $50.0/45.3$ & $35.4/34.2$ & $78.6/77.2$ & $77.7/67.1$ & $73.5/70.5$ & $47.4/41.3$ & $67.7/64.1$ \\
Qwen-SFT & \textcolor{gray}{{arXiv'25}} 
& $88.9/86.6$ & $87.4/87.5$ & $66.8/66.3$ & $47.3/44.2$ & $73.4/72.8$ & $81.3/73.9$ & $67.4/65.0$ & $53.2/46.1$ & $70.7/67.8$ \\
Q-Align & \textcolor{gray}{{ICML'24}} 
& $94.1/94.0$ & $88.6/88.7$ & $67.4/68.4$ & $40.3/41.9$ & $85.3/86.0$ & $77.2/73.5$ & $78.5/73.7$ & $55.4/48.3$ & $73.4/71.8$ \\
DeQA-Score & \textcolor{gray}{{CVPR'25}} 
& $\underline{95.3}/\underline{94.1}$ & $89.5/89.6$ & $69.4/68.7$ & $47.2/47.8$ & $\underline{89.2}/\underline{87.9}$ & $80.9/72.9$ & $\underline{78.7}/\underline{74.4}$ & $58.9/\underline{50.1}$ & $\underline{76.1}/\underline{73.2}$ \\
Q-insight & \textcolor{gray}{{NeurIPS'25}} 
& $91.8/89.5$ & $\underline{90.3}/\underline{89.9}$ & $70.2/70.2$ & $46.1/45.3$ & $87.0/83.9$ & $\underline{81.6}/\mathbf{76.6}$ & $68.5/64.0$ & $\underline{59.7}/48.6$ & $74.5/71.0$ \\
VisualQuality-R1 & \textcolor{gray}{{NeurIPS'25}} 
& $92.1/90.8$ & $87.7/87.8$ & $\underline{72.3}/\underline{71.9}$ & $\underline{53.1}/\underline{48.6}$ & $85.6/82.7$ & $\mathbf{81.7}/\underline{76.0}$ & $76.8/70.7$ & $57.2/47.1$ & $75.8/72.0$ \\
\rowcolor{gray!20}\textbf{Ours}  & --
& $\mathbf{95.9}/\mathbf{95.1}$ & $\mathbf{90.4}/\mathbf{90.3}$ & $\mathbf{77.9}/\mathbf{77.5}$ & $\mathbf{55.1}/\mathbf{55.0}$ & $\mathbf{90.9}/\mathbf{88.0}$ & $80.7/75.2$ & $\mathbf{84.5}/\mathbf{80.6}$ & $\mathbf{64.0}/\mathbf{51.3}$ & $\mathbf{80.0}/\mathbf{76.2}$ \\
\bottomrule
\end{tabular}}
\label{tab:main_single}
\end{table*}

\begin{table*}[t]
\centering
\caption{\textbf{PLCC / SRCC comparison between our Q-Hawkeye and multi-dataset training baselines.} Q-Hawkeye is trained only on \textcolor{t2e_blue}{KonIQ}, while others use multiple datasets. The best and second-best results per dataset are marked in \textbf{bold} and \underline{underline}.}
\vspace{0.1cm}
\setlength{\tabcolsep}{3pt}
\renewcommand{\arraystretch}{1.3}
\resizebox{\linewidth}{!}{
\begin{tabular}{c|c|cccccccc|c}
\toprule
\textbf{Method} & \textbf{Training Dataset} & \textbf{KonIQ} & \textbf{SPAQ} & \textbf{KADID} & \textbf{PIPAL} & \textbf{LIVE-Wild} & \textbf{AGIQA-3K} & \textbf{CSIQ} & \textbf{FLIVE} & \textbf{Avg} \\
\midrule
\multirow{5}{*}{DeQA-Score} & \multirow{2}{*}{\textcolor{t2e_blue}{KonIQ}, \textcolor{t2e_green}{SPAQ}}
& \multirow{2}{*}{$95.3/94.3$} & \multirow{2}{*}{$93.6/93.3$} & \multirow{2}{*}{$72.4/71.9$} & \multirow{2}{*}{$46.8/47.4$} & \multirow{2}{*}{$\underline{90.2}/\mathbf{88.8}$} & \multirow{2}{*}{$\underline{81.0}/73.8$} & \multirow{2}{*}{$83.6/78.5$} & \multirow{2}{*}{$45.8/39.0$} & \multirow{2}{*}{$76.1/73.4$} \\
& & & & & & & & & & \\
\cline{2-11}
& \textcolor{t2e_blue}{KonIQ}, \textcolor{t2e_green}{SPAQ},
& \multirow{2}{*}{$95.7/94.4$} & \multirow{2}{*}{$\mathbf{93.8}/\mathbf{93.4}$} & \multirow{2}{*}{$\underline{95.5}/\underline{95.3}$} & \multirow{2}{*}{$49.5/49.6$} & \multirow{2}{*}{$90.0/\underline{88.7}$} & \multirow{2}{*}{$80.8/74.5$} & \multirow{2}{*}{$\mathbf{90.0}/\mathbf{85.7}$} & \multirow{2}{*}{$48.2/42.1$} & \multirow{2}{*}{$\underline{80.4}/\underline{77.9}$} \\
& \textcolor{t2e_red}{KADID} & & & & & & & & & \\
\cline{2-11}
& \textcolor{t2e_blue}{KonIQ}, \textcolor{t2e_green}{SPAQ},
& \multirow{2}{*}{$\underline{95.8}/\underline{94.6}$} & \multirow{2}{*}{$\underline{93.2}/\underline{92.9}$} & \multirow{2}{*}{$\mathbf{96.3}/\mathbf{96.1}$} & \multirow{2}{*}{$\mathbf{72.4}/\mathbf{69.0}$} & \multirow{2}{*}{$87.7/85.7$} & \multirow{2}{*}{$77.0/73.5$} & \multirow{2}{*}{$86.3/80.7$} & \multirow{2}{*}{$58.9/\underline{50.1}$} & \multirow{2}{*}{$\mathbf{83.5}/\mathbf{80.3}$} \\
& \textcolor{t2e_red}{KADID}, \textcolor{t2e_gray}{PIPAL} & & & & & & & & & \\
\cline{1-11}
\multirow{2}{*}{VisualQuality-R1} & \multirow{2}{*}{\textcolor{t2e_red}{KADID}, \textcolor{t2e_green}{SPAQ}}
& \multirow{2}{*}{$89.0/87.4$} & \multirow{2}{*}{$91.7/91.3$} & \multirow{2}{*}{$86.7/86.8$} & \multirow{2}{*}{$43.7/43.4$} & \multirow{2}{*}{$85.6/82.7$} & \multirow{2}{*}{$80.0/73.3$} & \multirow{2}{*}{$85.5/79.7$} & \multirow{2}{*}{$53.1/47.0$} & \multirow{2}{*}{$76.9/74.0$} \\
& & & & & & & & & & \\
\cline{1-11}
\multirow{2}{*}{Q-insight} & \multirow{2}{*}{\textcolor{t2e_blue}{KonIQ}, \textcolor{t2e_purple}{KADIS}}
& \multirow{2}{*}{$93.3/91.6$} & \multirow{2}{*}{$90.7/90.5$} & \multirow{2}{*}{$74.2/73.6$} & \multirow{2}{*}{$48.6/47.4$} & \multirow{2}{*}{$89.3/86.5$} & \multirow{2}{*}{$\textbf{81.6}/\mathbf{76.6}$} & \multirow{2}{*}{$\underline{87.0}/\underline{82.4}$} & \multirow{2}{*}{$\underline{62.0}/49.9$} & \multirow{2}{*}{$78.3/74.8$} \\
& & & & & & & & & & \\
\cline{1-11}
\rowcolor{gray!20}
Ours & \textcolor{t2e_blue}{KonIQ}
& $\mathbf{95.9}/\mathbf{95.1}$ & $90.4/90.3$ & $77.9/77.5$ & $\underline{55.1}/\underline{55.0}$ & $\mathbf{90.9}/88.0$ & $80.7/75.2$ & $84.5/80.6$ & $\mathbf{64.0}/\mathbf{51.3}$ & $80.0/76.2$ \\
\bottomrule
\end{tabular}}
\label{tab:main_multi}
\vspace{-6pt} 
\end{table*}

\subsection{Main Result}
\textbf{Comparison with single-dataset training methods.} As shown in Tab.~\ref{tab:main_single}, when \textbf{all methods are trained only on KonIQ~\cite{KonIQ} dataset}, Q-Hawkeye achieves the best average PLCC/SRCC across eight datasets, surpassing both traditional CNN/Transformer-based IQA like MUSIQ~\cite{MUSIQ}, CLIP-IQA+~\cite{CLIP-IQA}, and recent MLLM-based methods such as Q-Align~\cite{Q-Align}, DeQA-Score~\cite{DeQA-Score}, Q-Insight~\cite{Q-insight}, and VisualQuality-R1~\cite{Visionary-R1}. In particular, our method not only exceeds the strongest baselines on the in-domain KonIQ dataset, but it also exhibits clear advantages in cross-dataset generalization. For example, compared with VisualQuality-R1,  Q-Hawkeye improves PLCC from 72.3 and 53.1 to \textbf{77.9} and \textbf{55.1}, improves SRCC from 71.9 and 48.6 to \textbf{77.5} and \textbf{55.0} on KADID~\cite{KIDID-10K} and PIPAL~\cite{gu2020pipallargescaleimagequality} datasets, respectively. Q-Hawkeye also achieves the best results on CSIQ (\textbf{84.5/80.6}) and FLIVE~\cite{FLIVE} (\textbf{64.0/51.3}). This suggests that  Q-Hawkeye learns quality judgments that better transfer across different contents and degradation types.

\begin{table*}[h]
\centering
\caption{\textbf{Ablation studies} on proposed modules with PLCC / SRCC metrics. Models are trained on KonIQ dataset.}
\label{tab:ablation}
\resizebox{\textwidth}{!}{%
\begin{tabular}{c|c|cc|c|ccccccc|c}
\toprule
\multirow{2}{*}{\#} & \multirow{2}{*}{\makecell{Uncertainty\\-Aware}} & \multicolumn{2}{c|}{Perception-Aware} & In-dist. & \multicolumn{7}{c|}{Out-of-distribution} & \multirow{2}{*}{Avg} \\
\cmidrule(lr){3-4} \cmidrule(lr){5-5} \cmidrule(lr){6-12}
 &  & KL term & Entropy term & KonIQ & SPAQ & KADID & PIPAL & LIVE-Wild & AGIQA-3K & CSIQ & FLIVE &  \\
\midrule
1 & \xmark & \xmark & \xmark & 90.8/88.5 & 85.6/85.2 & 69.2/68.8 & 44.5/42.3 & 86.1/82.8 & 81.3/76.8 & 78.2/74.5 & 62.2/59.2 & 74.8/72.3 \\
2 & \cmark & \xmark & \xmark & 92.5/90.8 & 87.2/86.8 & 72.5/72.1 & 47.8/46.2 & 87.8/84.5 & \textbf{81.9}/\textbf{78.5} & 80.0/76.2 & 68.9/55.4 & 77.5/73.8 \\
3 & \cmark & \cmark & \xmark & 94.6/93.8 & 89.5/89.3 & 76.0/75.6 & 52.5/52.2 & 89.6/86.8 & 81.0/75.8 & 82.5/78.8 & 63.0/50.2 & 78.6/75.3 \\
4 & \cmark & \xmark & \cmark & 93.2/91.8 & 87.8/87.5 & 73.8/73.4 & 49.5/49.0 & 87.5/84.2 & 81.5/77.6 & 79.8/76.0 & 61.2/47.8 & 76.9/73.4 \\
5 & \xmark & \cmark & \cmark & 94.5/93.6 & 89.0/88.8 & 76.2/75.8 & 53.2/52.8 & 90.9/87.2 & 79.8/74.5 & \textbf{84.8}/\textbf{81.0} & 63.2/50.6 & 78.9/75.5 \\
\rowcolor{gray!20}
6 & \cmark & \cmark & \cmark & \textbf{95.9}/\textbf{95.1} & \textbf{90.4}/\textbf{90.3} & \textbf{77.9}/\textbf{77.5} & \textbf{55.1}/\textbf{55.0} & \textbf{91.0}/\textbf{88.0} & 80.7/75.2 & 84.5/80.6 & \textbf{64.0}/\textbf{51.3} & \textbf{80.0}/\textbf{76.6} \\
\bottomrule
\end{tabular}%
}
\vspace{-6pt} 
\end{table*}
\begin{table}[h]
\centering
\caption{Ablation study on the value of $\tau$.}
\vspace{-7pt} 
\label{tab:tau}
\setlength{\tabcolsep}{3pt}
\small
\begin{tabular}{l|c|c|c|c|c}
\toprule
Dataset / $\tau$ & 0.1 & 0.2 & 0.3 & 0.4 & 0.5 \\
\midrule
KonIQ & 92.3/91.5 & \textbf{95.9/95.1} & 94.8/94.0 & 93.5/92.7 & 91.8/91.0 \\
SPAQ  & 86.9/86.8 & \textbf{90.4/90.3} & 89.1/89.0 & 87.8/87.7 & 86.2/86.1 \\
KADID & 74.2/73.8 & \textbf{77.9/77.5} & 76.5/76.1 & 75.3/74.9 & 73.8/73.4 \\
PIPAL & 51.6/51.5 & \textbf{55.1/55.0} & 53.9/53.8 & 52.7/52.6 & 51.2/51.1 \\
LiveW & \textbf{91.5/88.6} & 90.9/88.0 & 89.7/86.8 & 88.4/85.5 & 87.0/84.1 \\
AGIQA & 77.1/71.6 & 80.7/75.2 & \textbf{81.2/75.8} & 79.3/73.8 & 77.5/72.0 \\
CSIQ  & 80.8/76.9 & \textbf{84.5/80.6} & 83.2/79.3 & 81.9/78.0 & 80.3/76.4 \\
FLIVE & \textbf{64.8/52.1} & 64.0/51.3 & 62.6/49.9 & 61.2/48.5 & 59.7/47.0 \\
\midrule
\rowcolor{gray!20}
Avg   & 77.4/74.1 & \textbf{80.0/76.6} & 78.9/75.6 & 77.5/74.2 & 75.9/72.6 \\
\bottomrule
\end{tabular}
\end{table}
\begin{table}[h]
\centering
\caption{Ablation study on the value of $\gamma$.}
\vspace{-7pt} 
\label{tab:gamma}
\setlength{\tabcolsep}{3pt}
\small
\begin{tabular}{l|c|c|c|c|c}
\toprule
Dataset / $\gamma$  & 5e-3 & 1e-3 & 5e-4 & 1e-4 & 5e-5 \\
\midrule
KonIQ & 86.5/85.7 & 94.6/93.8 & \textbf{95.9/95.1} & 93.2/92.4 & 91.8/91.0 \\
SPAQ  & 81.2/81.1 & 89.1/89.0 & \textbf{90.4/90.3} & 87.5/87.4 & 86.3/86.2 \\
KADID & 68.7/68.3 & 76.5/76.1 & \textbf{77.9/77.5} & 74.8/74.4 & 73.6/73.2 \\
PIPAL & 46.3/46.2 & 53.8/53.7 & \textbf{55.1/55.0} & 52.4/52.3 & 51.2/51.1 \\
LiveW & 82.4/79.5 & \textbf{91.3/88.4} & 90.9/88.0 & 88.6/85.7 & 87.4/84.5 \\
AGIQA & 71.5/66.0 & 79.2/73.7 & 80.7/75.2 & \textbf{81.1/75.6} & 79.8/74.3 \\
CSIQ  & 75.3/71.4 & 83.1/79.2 & \textbf{84.5/80.6} & 82.3/78.4 & 81.0/77.1 \\
FLIVE & 65.7/53.0 & \textbf{64.8/52.1} & 64.0/51.3 & 62.8/50.1 & 61.5/48.8 \\
\midrule
\rowcolor{gray!20}
Avg   & 72.2/68.9 & 79.0/75.8 & \textbf{80.0/76.6} & 77.8/74.5 & 76.6/73.3 \\
\bottomrule
\end{tabular}
\vspace{-12pt} 
\end{table}

\textbf{Comparison with multi-dataset training methods.} Tab.~\ref{tab:main_multi} further compares Q-Hawkeye with methods trained on multiple datasets.  Despite using only the KonIQ dataset for training, Q-Hawkeye achieves competitive or better average performance than RL-based multi-dataset methods. For example, VisualQuality-R1 and Q-Insight achieve $75.6/74.0$ and $78.3/74.8$ in average of PLCC/SRCC on eight datasets, respectively, while Q-Hawkeye reaches $\textbf{80.0/76.2}$ with single-dataset training, and comes close to the DeQA-Score trained on three and four datasets.  On KonIQ, LiveW, and FLIVE datasets, our single-dataset training model even outperforms these multi-dataset baselines, indicating that designing a reliable learning signal is critical. By jointly leveraging uncertainty-aware dynamic optimization and perception-aware constraints, Q-Hawkeye attains strong, data-efficient generalization across diverse IQA benchmarks.

\subsection{Ablation Studies}
\textbf{Effectiveness of proposed modules.} We conduct ablation studies on the proposed modules of our method, as summarized in Tab.~\ref{tab:ablation}. First, from rows 1, 2, and 5, we observe that enabling either uncertainty-aware weighting or perception-aware optimization consistently improves model performance over the GRPO baseline, showing that both components are independently beneficial. Second, rows 3, 4, and 6 reveal the role of the two perception terms that adding the KL term on top of uncertainty-aware training already brings gains, while using only the entropy term leads to a slight performance degradation; the best performance appears when KL and entropy are combined in row 6, indicating that entropy mainly acts as a stabilizer that works best together with the KL-guided perception constraint. Finally, comparing rows 1, 2, and 6, we observe a clear performance lift from baseline to uncertainty-aware only, and further with both uncertainty-aware and perception-aware modules, confirming that the two designs are complementary in improving robustness and cross-dataset generalization.

\begin{figure}[H]
  \centering
\vspace{-8pt} 
\includegraphics[width=1.0\linewidth]{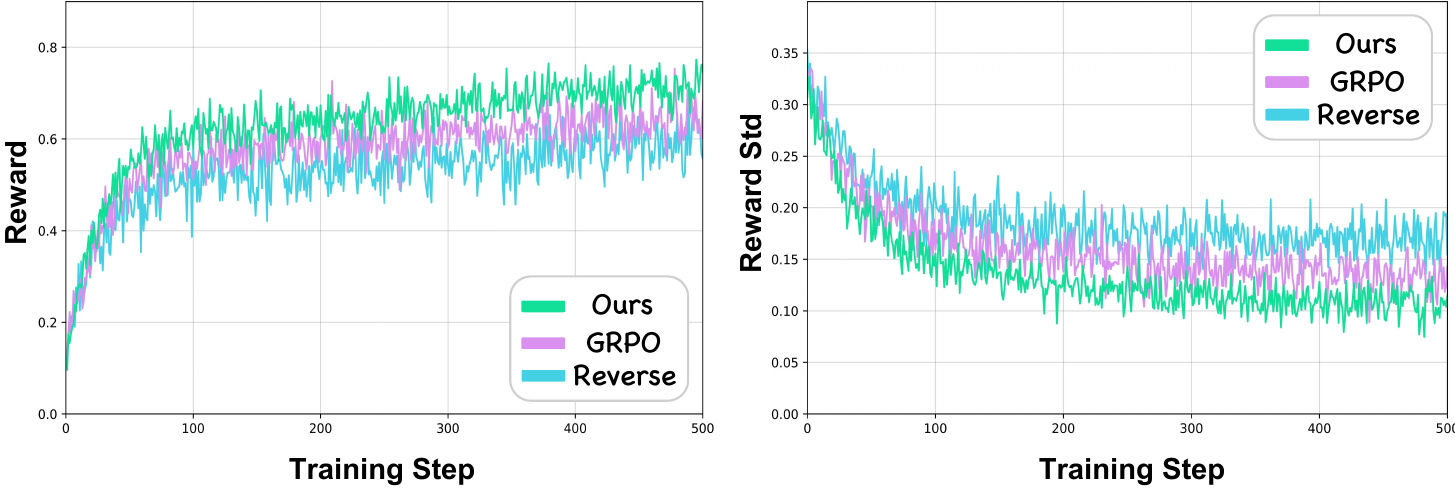}
\caption{Training dynamics of uncertainty-aware optimization.Average reward (left) and reward std (right) for vanilla GRPO, our uncertainty-aware weighting, and a reverse weighting baseline.}
\label{fig:Reward_Curve}
\vspace{-16pt} 
\end{figure}

We further study the effect of uncertainty-aware dynamic optimization by monitoring the reward and its standard deviation during training (Fig.~\ref{fig:Reward_Curve}). Compared with vanilla GRPO, our uncertainty-aware weighting achieves consistently higher rewards and lower reward variance, indicating a more stable optimization process. In contrast, the `Reverse' variant, which mistakenly upweights high-uncertainty samples and downweights low-uncertainty ones, shows both lower final rewards and clearly larger reward fluctuations. These trends confirm that correctly down-weighting ambiguous samples is crucial for reliable and stable policy learning. \textbf{Ablation studies on Perception-Aware Optimization module components are provided in the Appendix~\ref{sec:app_perception}.}

\textbf{Ablation studies on parameters $\tau$ and $\gamma$}.
$\tau$ controls the strength of uncertainty-based reweighting, and $\gamma$ weights the implicit perception loss. Tab.~\ref{tab:tau} and Tab.~\ref{tab:gamma} report the PLCC/SRCC results on all eight datasets under different choices of the uncertainty temperature $\tau$ and the perception-loss weight $\gamma$, respectively. As shown in both tables, Q-Hawkeye maintains robust performance across a wide range of settings, with only minor performance variations. Based on these observations, we adopt $\tau = 0.2$ and $\gamma = 5 \times 10^{-4}$ as the default configuration in our main experiments.

\begin{figure}[H]
  \centering
\vspace{-12pt} 
\includegraphics[width=0.80\linewidth]{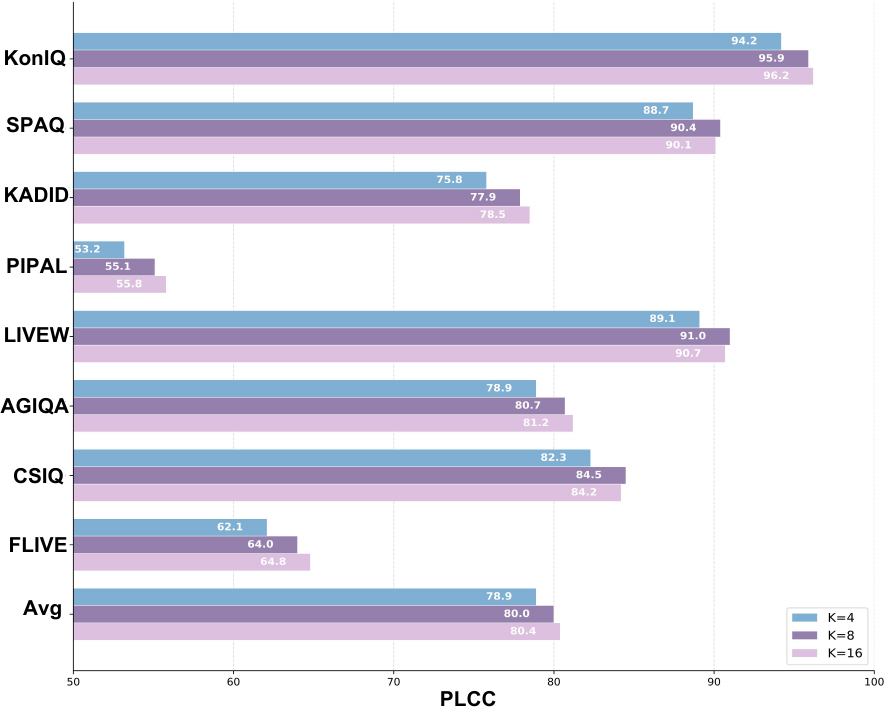}
\caption{PLCC performance on each dataset when varying the number of generated responses $K$.}
 \label{fig:K_Size}
\label{K_Size}
\vspace{-12pt}  
\end{figure}

\textbf{Ablation studies on the number of rollouts $K$.} We also investigate how the number of generated responses per image $K$ affects our method by trying $ K = 4$, $8$, and $16$ while keeping all other settings fixed. As shown in Fig.~\ref{K_Size}, the performance remains stable across different $K$ values: increasing $K$ from 4 to 8 brings a small but consistent gain, while further increasing to 16 yields only marginal improvement. Considering both computation cost and accuracy, we finally set $K$ = 8 as the default configuration in our main experiments. \textbf{More ablation results on other hyperparameters are provided in the Appendix~\ref{sec:app_ablation}.}

\section{Conclusion}
In this paper, we propose Q-Hawkeye, a GRPO-based framework for reliable visual policy optimization in image quality assessment. Built on Qwen2.5-VL-7B, Q-Hawkeye reshapes the RL learning signal from two complementary perspectives: an Uncertainty-Aware Dynamic Optimization strategy that adaptively reweights per image updates based on score variance across rollouts, and a Perception-Aware Optimization module that enforces consistent distributional differences between original and degraded images via an implicit perception loss with double entropy regularization. These components jointly encourage the policy to focus on stable, informative training samples while grounding its judgments in genuine visual evidence rather than spurious textual priors. Extensive experiments on eight IQA benchmarks further demonstrate the effectiveness of the proposed modules,  and show that Q-Hawkeye consistently outperforms existing state-of-the-art methods in both single- and multi-dataset settings, with clear gains in average PLCC/SRCC metrics and improves the model's robustness on challenging out-of-distribution distortions.


\section{Impact Statement}
\label{sec:impact_state}
This work aims to advance machine learning by improving the reliability of MLLM-based no-reference image quality assessment through uncertainty-aware optimization and perception-aware constraints. More reliable IQA may benefit practical systems such as image/video enhancement, compression, and quality control for AI-generated content.

\bibliography{example_paper}
\bibliographystyle{icml2026}

\newpage
\appendix
\onecolumn

\begin{center}
{\LARGE \textbf{————Appendix————}}
\end{center}

\vspace{1.5em}
\noindent\textbf{\large Contents}
\vspace{1em}

{
\hypersetup{linkcolor=black}
\noindent\makebox[1.5em][l]{\textbf{A}}\textbf{Details of Q-Hawkeye Compared to Existing Methods}\dotfill\pageref{sec:app_details}

\vspace{0.3em}
\hspace{1.5em}\makebox[2em][l]{A.1}Compared with MLLM-based IQA Methods \dotfill\pageref{sec:app_mllm}

\vspace{0.3em}
\hspace{1.5em}\makebox[2em][l]{A.2}Compared with Existing RL-based Methods \dotfill\pageref{sec:app_rl}

\vspace{0.8em}
\noindent\makebox[1.5em][l]{\textbf{B}}\textbf{More Results of Ablation Studies}\dotfill\pageref{sec:app_ablation}

\vspace{0.3em}
\hspace{1.5em}\makebox[2em][l]{B.1}Ablation study on $\alpha$ and $\eta_1,\eta_2$ \dotfill\pageref{sec:app_alpha}

\vspace{0.3em}
\hspace{1.5em}\makebox[2em][l]{B.2}Ablation study on Batch Size \dotfill\pageref{sec:app_batch}

\vspace{0.3em}
\hspace{1.5em}\makebox[2em][l]{B.3}Effectiveness of Perception-Aware Optimization \dotfill\pageref{sec:app_perception}

\vspace{0.8em}
\noindent\makebox[1.5em][l]{\textbf{C}}\textbf{Complexity Analysis}\dotfill\pageref{sec:app_complexity}

\vspace{0.8em}
\noindent\makebox[1.5em][l]{\textbf{D}}\textbf{Detail of Evaluated Datasets}\dotfill\pageref{sec:app_datasets}

\vspace{0.8em}
\noindent\makebox[1.5em][l]{\textbf{E}}\textbf{Details of Evaluation Metrics}\dotfill\pageref{sec:app_metrics}

\vspace{0.8em}
\noindent\makebox[1.5em][l]{\textbf{F}}\textbf{Details of Degraded Image Construction}\dotfill\pageref{sec:app_degraded}

\vspace{0.8em}
\noindent\makebox[1.5em][l]{\textbf{G}}\textbf{Details of Hyperparameters}\dotfill\pageref{app:hyperparameters}

\vspace{0.8em}
\noindent\makebox[1.5em][l]{\textbf{H}}\textbf{More Visualization Examples}\dotfill\pageref{sec:app_vis}


\vspace{0.5em}
\noindent\hrulefill
\vspace{1em}

\section{Details of Q-Hawkeye Compared to Existing Methods}
\label{sec:app_details}
\subsection{Compared with MLLM-based IQA Methods.}
\label{sec:app_mllm}
Recent MLLM-based IQA methods leverage MLLMs’ strong vision--language alignment and instruction-following abilities to generate text-grounded quality rationales alongside scalar scores, thereby improving generalization beyond purely discriminative regressors. Among them, \textbf{Q-Align}\cite{Q-Align} formulates quality prediction as selecting from discrete quality levels (and can serve as an alignment signal), but such discretization may blur fine-grained MOS differences and does not directly address optimization reliability. \textbf{DeQA-Score}\cite{DeQA-Score} improves supervised training by modeling scores in a distributional manner to provide denser regression supervision; however, it remains SFT-driven and does not account for the fact that prediction stability can vary substantially across samples, making it prone to noisy updates when the model has not formed reliable judgments. Moving to RL-based post-training, \textbf{Q-Insight}\cite{Q-insight} applies GRPO with verifiable rewards to jointly optimize score prediction and distortion awareness, yet its update strength is still largely uniform across samples under GRPO-style optimization, and perceptual reliability is mainly encouraged through task design rather than explicitly enforcing that the policy reacts to changes in visual evidence. \textbf{VisualQuality-R1}\cite{Visionary-R1} further reformulates NR-IQA as a pairwise ranking problem and uses multiple rollouts to induce reasoning-aware ranking behavior, but it similarly lacks an explicit mechanism to suppress the gradient impact of high-uncertainty instances and to guarantee that aligned scores/rankings are truly grounded in visual cues instead of dataset regularities or language priors.

In contrast, Q-Hawkeye treats reliability as a first-class training objective and differs from prior MLLM-based IQA methods in two complementary ways. First, we explicitly quantify sample-wise predictive uncertainty using the variance of rollout scores and inject it into GRPO via uncertainty-aware advantage reweighting, preventing unstable samples from contributing disproportionately noisy gradients. Second, we strengthen perceptual grounding by constructing pristine--degraded pairs and introducing an Implicit Perception Loss that enforces separable policy output distributions under different visual conditions, together with a regularization that suppresses degenerate high-entropy expansion. This unified design enables Q-Hawkeye to optimize not only for score alignment but also for stable learning dynamics and visually grounded quality reasoning, improving robustness and cross-dataset generalization without requiring distortion labels as mandatory supervision.

\subsection{Compared with Existing RL-based Methods.}
\label{sec:app_rl}
\textbf{Dynamic Reinforcement Learning.}
Recent GRPO variants explore exploiting uncertainty signals to modulate post-training. For instance, SEED-GRPO~\cite{SEED-GRPO} uses semantic entropy as a prompt-level uncertainty to scale update magnitudes, while MAPO~\cite{MAPO} addresses advantage reversion by reweighting advantage formulations according to trajectory certainty under discrete success signals; GRPO-CARE~\cite{GRPO-CARE} instead improves multimodal reliability via a consistency-aware bonus that encourages coherent reasoning-to-answer behavior. Different from prior works, for the IQA task with continuous and subjective labels, Q-Hawkeye adopts a task-native signal---rollout score variance across $K$ responses for the same image, and leverages it with GRPO through sample-wise advantage rescaling, where a single weight $w(u)$ directly controls per-sample gradient variance while preserving GRPO’s within-group relative learning. For fairness, we also implement a SEED-style baseline that computes uncertainty via semantic entropy and applies the same dynamic advantage reweighting. Table~\ref{tab:Entropy} shows that it underperforms our score-variance-based scaling. We attribute this to semantic entropy, which mainly reflects variation in textual rationales and noisy semantic grouping, which can inflate uncertainty without corresponding score ambiguity, whereas score variance is bounded and directly aligned with the scalar prediction target, yielding lower variance and a stronger signal for advantage-strength control.

\begin{table}[h]
\vspace{-4pt}  
\centering
\caption{Comparison between a semantic-entropy uncertainty scaling baseline and our strategy. All models are trained on the KonIQ-10k dataset with same backbone, and report PLCC/SRCC ($\uparrow$) on eight datasets, and Avg denotes the average performance across datasets.}
\resizebox{\linewidth}{!}{
\large
\begin{tabular}{lccccccccc}
\toprule
Model & KonIQ & SPAQ & KADID & PIPAL & LIVE-Wild & AGIQA-3K & CSIQ & FLIVE & Avg $\uparrow$ \\
\midrule
Entropy & 93.2/92.5 & 88.1/87.8 & 74.5/74.0 & 52.3/51.8 & 88.5/85.2 & 78.2/72.5 & 81.2/77.8 & 44.8/37.6 & 75.1/72.4 \\
\textbf{Ours} & \textbf{95.9/95.1} & \textbf{90.4/90.3} & \textbf{77.9/77.5} & \textbf{55.1/55.0} & \textbf{91.0/88.0} & \textbf{80.7/75.2} & \textbf{84.5/80.6} & \textbf{64.0/51.3} & \textbf{80.0/76.6} \\
\midrule
\rowcolor{gray!20}
Improve($\Delta$) & \textcolor{green!50!black}{+2.7/+2.6} & \textcolor{green!50!black}{+2.3/+2.5} & \textcolor{green!50!black}{+3.4/+3.5} & \textcolor{green!50!black}{+2.8/+3.2} & \textcolor{green!50!black}{+2.5/+2.8} & \textcolor{green!50!black}{+2.5/+2.7} & \textcolor{green!50!black}{+3.3/+2.8} & \textcolor{green!50!black}{+19.2/+13.7} & \textcolor{green!50!black}{+4.9/+4.2} \\
\bottomrule
\end{tabular}
\label{tab:Entropy}
}
\vspace{-5pt}  
\end{table}

\begin{table*}[h]
\centering
\caption{ We compare PAPO-style masking strategy with our original--degraded pairing srtategy, with and without Uncertainty-Aware advantage reweighting. All models are trained on KonIQ and evaluated using PLCC/SRCC on KonIQ and seven OOD benchmarks; Avg denotes the mean across datasets.}
\label{tab:ablation_papo}
\resizebox{\textwidth}{!}{%
\begin{tabular}{c|c|cc|c|ccccccc|c}
\toprule
\multirow{2}{*}{\#} & \multirow{2}{*}{\makecell{Uncertainty\\-Aware}} & \multicolumn{2}{c|}{Perception-Aware} & In-dist. & \multicolumn{7}{c|}{Out-of-distribution} & \multirow{2}{*}{Avg} \\
\cmidrule(lr){3-4} \cmidrule(lr){5-5} \cmidrule(lr){6-12}
 &  & Mask & Degrade & KonIQ & SPAQ & KADID & PIPAL & LIVE-Wild & AGIQA-3K & CSIQ & FLIVE &  \\
\midrule
1 & \xmark & \cmark & \xmark & 89.6/87.0 & 84.0/83.6 & 67.6/67.0 & 42.2/40.6 & 84.6/81.2 & 77.2/72.6 & 75.8/72.0 & 45.4/42.0 & 70.8/68.5 \\
\rowcolor{gray!10}
2 & \xmark & \xmark & \cmark & 94.5/93.6 & 89.0/88.8 & 76.2/75.8 & 53.2/52.8 & 90.9/87.2 & 79.8/74.5 & \textbf{84.8}/\textbf{81.0} & 63.2/50.6 & 78.9/75.5 \\
\rowcolor{gray!20}
\multicolumn{4}{c|}{Improve($\Delta$)} & \textcolor{green!50!black}{+4.9/+6.6} & \textcolor{green!50!black}{+5.0/+5.2} & \textcolor{green!50!black}{+8.6/+8.8} & \textcolor{green!50!black}{+11.0/+12.2} & \textcolor{green!50!black}{+6.3/+6.0} & \textcolor{green!50!black}{+2.6/+1.9} & \textcolor{green!50!black}{+9.0/+9.0} & \textcolor{green!50!black}{+17.8/+8.6} & \textcolor{green!50!black}{+8.1/+7.0} \\
\midrule
3 & \cmark & \cmark & \xmark & 90.2/87.8 & 84.8/84.5 & 68.5/68.0 & 43.2/41.5 & 85.5/82.0 & 78.0/73.5 & 76.8/72.8 & 45.8/42.7 & 71.6/69.1 \\
\rowcolor{gray!10}
4 & \cmark & \xmark & \cmark & \textbf{95.9}/\textbf{95.1} & \textbf{90.4}/\textbf{90.3} & \textbf{77.9}/\textbf{77.5} & \textbf{55.1}/\textbf{55.0} & \textbf{91.0}/\textbf{88.0} & \textbf{80.7}/\textbf{75.2} & 84.5/80.6 & \textbf{64.0}/\textbf{51.3} & \textbf{80.0}/\textbf{76.6} \\
\rowcolor{gray!20}
\multicolumn{4}{c|}{Improve($\Delta$)} & \textcolor{green!50!black}{+5.7/+7.3} & \textcolor{green!50!black}{+5.6/+5.8} & \textcolor{green!50!black}{+9.4/+9.5} & \textcolor{green!50!black}{+11.9/+13.5} & \textcolor{green!50!black}{+5.5/+6.0} & \textcolor{green!50!black}{+2.7/+1.7} & \textcolor{green!50!black}{+7.7/+7.8} & \textcolor{green!50!black}{+18.2/+8.6} & \textcolor{green!50!black}{+8.4/+7.5} \\
\bottomrule
\end{tabular}%
}
\vspace{-5pt} 
\end{table*}

\textbf{Perception-Aware Reinforcement Learning.} Recent perception-aware RL methods for MLLMs typically introduce explicit objectives to discourage ``reasoning without seeing'' by enforcing stronger dependence on visual evidence. For instance, PAPO~\cite{PAPO} contrasts the policy conditioned on the original image with that on a masked image; Perception-R1~\cite{Perception-R1} studies rule-based RL for perception policy learning and shows that perceptual perplexity and reward design largely govern RL effectiveness across perception tasks (e.g., grounding, counting, OCR, detection); and Visionary-R1~\cite{Visionary-R1} mitigates shortcut learning in GRPO by enforcing an ``interpret-before-reason'' caption--reason--answer format with additional caption supervision. In contrast, Q-Hawkeye targets IQA where supervision is a continuous and subjective quality score. Instead of masking away semantic content or requiring explicit captioning, we construct  original--degraded pairs that preserve scene semantics while isolating quality-related evidence via an effective-contrast filtering strategy, and maximize the distributional discrepancy between $\pi_\theta(\cdot\!\mid\! I,q)$ and $\pi_\theta(\cdot\!\mid\! I^{\mathrm{deg}},q)$ with a Perception Loss to explicitly force the policy to react to degradation cues. Tab.~\ref{tab:ablation_papo} provides a controlled comparison between the PAPO-style masking signal and our degradation-based design, with and without Uncertainty-Aware optimization: replacing masking with degradation yields consistently better IQA performance (rows 1 vs. rows 2; rows 3 vs. rows 4), and our Q-Hawkeye (rows 4) achieves the best in-distribution and cross-dataset results. We attribute this gap to the fact that heavy masking primarily removes semantic content rather than inducing graded quality degradations, so the resulting KL signal is often dominated by ``what is missing'' instead of ``how quality changes,'' weakening its alignment with scalar MOS regression and potentially harming score calibration. Similarly, perception-policy objectives designed for discrete perception tasks or shortcut-mitigation via captioning emphasize content correctness more than fine-grained distortion sensitivity.

\section{More Results of Ablation Studies}
\label{sec:app_ablation}
\subsection{Ablation study on $\alpha$ and $\eta_1,\eta_2$}
\label{sec:app_alpha}
The tolerance parameter $\alpha$ controls how sharply the accuracy reward decays with the absolute error between the predicted score and MOS, thereby balancing sensitivity to small prediction errors and robustness to annotation noise. The coefficients $\eta_1$ and $\eta_2$ weight the entropy regularization terms for the original and degraded images, regulating how strongly we penalize overly high-entropy policies under the two visual conditions. For simplicity, we keep $\eta_1$ and $\eta_2$ tied in all experiments and perform ablations over different values, as reported in Tab.~\ref{tab:eta}, while Tab.~\ref{tab:alpha} summarizes the results obtained with different choices of $\alpha$. As shown in these tables, Q-Hawkeye maintains consistently strong performance across a wide range of settings, indicating that our method is not overly sensitive to these hyperparameters. Based on this study, we set $\alpha = 0.3$ and $\eta_1 = \eta_2 = 1 \times 10^{-4}$ for all main experiments.

\begin{table}[h]
\centering
\caption{Ablation study on the value of $\eta_1$ and $\eta_2$.}
\label{tab:eta}
\setlength{\tabcolsep}{3pt}
\begin{tabular}{l|c|c|c|c|c}
\toprule
Dataset / $\eta_1$, $\eta_2$ & 5e-3 & 1e-3 & 5e-4 & 1e-4 & 5e-5 \\
\midrule
KonIQ & 92.1/91.3 & 94.5/93.7 & 95.2/94.4 & \textbf{95.9/95.1} & 94.8/94.0 \\
SPAQ & 86.8/86.5 & 88.9/88.6 & 89.6/89.5 & \textbf{90.4/90.3} & 89.2/89.0 \\
KADID & 74.2/73.8 & 76.3/75.9 & 77.1/76.7 & \textbf{77.9/77.5} & 76.5/76.1 \\
PIPAL & 51.5/51.2 & 53.6/53.4 & 54.3/54.2 & \textbf{55.1/55.0} & 53.8/53.6 \\
LiveW & 87.2/84.3 & 89.4/86.5 & \textbf{91.2/88.3} & 90.9/88.0 & 89.6/86.8 \\
AGIQA & 77.0/71.5 & 79.2/73.8 & 80.0/74.5 & 80.7/75.2 & \textbf{81.0/75.6} \\
CSIQ & 80.8/76.9 & 82.9/79.0 & 83.7/79.8 & \textbf{84.5/80.6} & 83.2/79.3 \\
FLIVE & 60.5/47.8 & 62.3/49.6 & 63.2/50.5 & \textbf{64.0/51.3} & 62.0/49.2 \\
\midrule
\rowcolor{gray!20}
Avg & 76.3/72.9 & 78.4/75.1 & 79.3/76.0 & \textbf{80.0/76.6} & 78.8/75.4 \\
\bottomrule
\end{tabular}
\end{table}

\begin{table}[h]
\centering
\caption{Ablation study on the value of $\alpha$.}
\label{tab:alpha}
\setlength{\tabcolsep}{3pt}
\begin{tabular}{l|c|c|c|c|c}
\toprule
Dataset / $\alpha$ & 0.1 & 0.2 & 0.3 & 0.4 & 0.5 \\
\midrule
KonIQ & 92.3/91.5 & 94.6/93.8 & \textbf{95.9/95.1} & 95.5/94.7 & 94.8/94.0 \\
SPAQ & 86.9/86.6 & 89.0/88.7 & \textbf{90.4/90.3} & 89.8/89.6 & 89.1/88.9 \\
KADID & 74.3/73.9 & 76.5/76.1 & \textbf{77.9/77.5} & 77.2/76.8 & 76.4/76.0 \\
PIPAL & 51.6/51.3 & 53.8/53.5 & \textbf{55.1/55.0} & 54.5/54.3 & 53.7/53.4 \\
LiveW & 87.5/84.6 & 89.6/86.7 & 90.9/88.0 & \textbf{91.3/88.4} & 90.2/87.3 \\
AGIQA & 77.2/71.8 & 79.5/74.0 & 80.7/75.2 & 80.2/74.8 & \textbf{81.0/75.5} \\
CSIQ & 81.0/77.1 & 83.2/79.3 & \textbf{84.5/80.6} & 84.0/80.1 & 83.3/79.4 \\
FLIVE & 60.8/48.0 & 62.5/49.8 & \textbf{64.0/51.3} & 63.0/50.2 & 61.8/49.0 \\
\midrule
\rowcolor{gray!20}
Avg & 76.5/73.1 & 78.6/75.2 & \textbf{80.0/76.6} & 79.4/76.1 & 78.8/75.4 \\
\bottomrule
\end{tabular}
\end{table}

\subsection{Ablation study on Batch Size}
\label{sec:app_batch}
We further investigate how the global batch size affects Q-Hawkeye by varying it while keeping all other settings fixed (see Fig.~\ref{fig:BS_Comparsion}). The results show that our method remains stable across different batch sizes, with a slight but consistent gain when increasing the batch size from 16 to 32, while a larger batch of 64 brings only marginal or even saturated improvement. Considering both optimization stability and computational efficiency, we adopt a batch size of 32 in all main experiments.

\begin{figure*}[h]
  \centering
\includegraphics[width=0.5\linewidth]{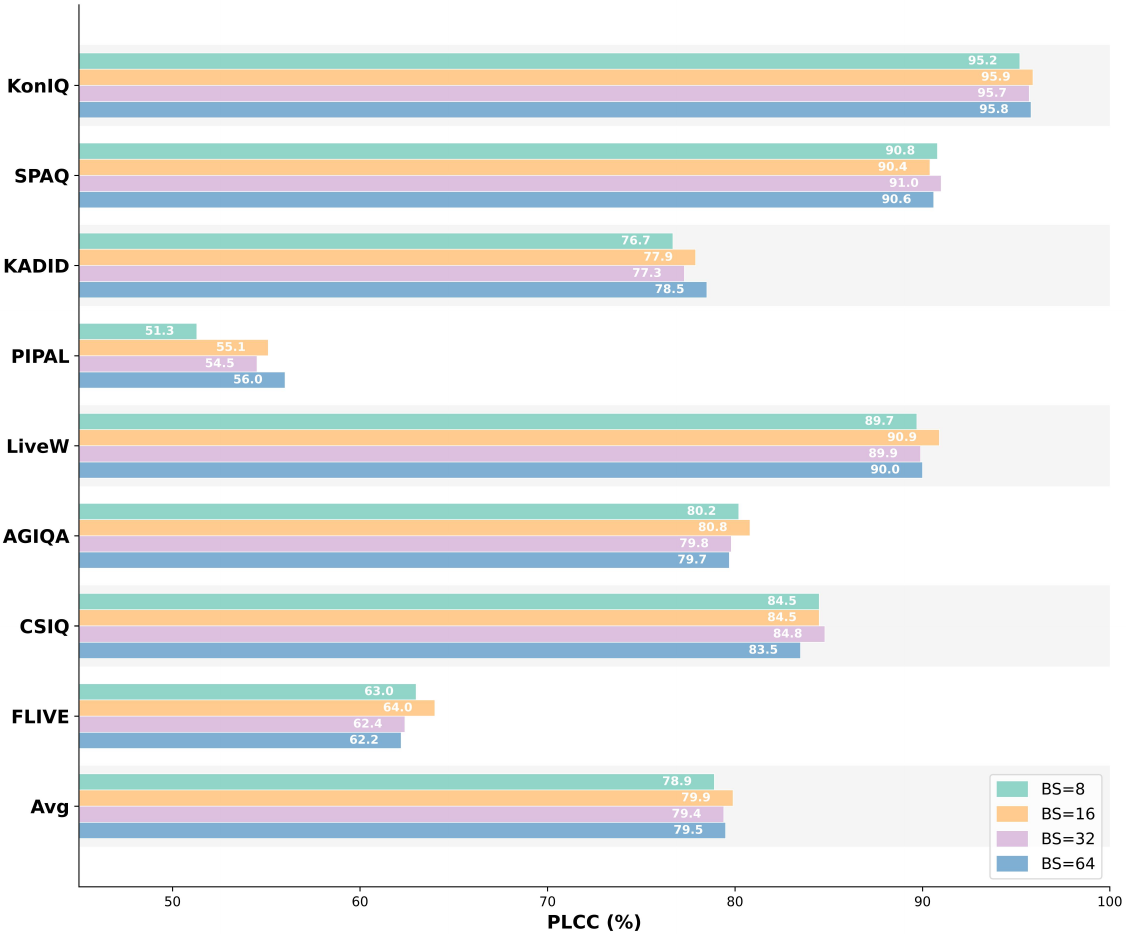}
\caption{PLCC performance on each dataset when varying the Batch Size.}
 \label{fig:BS_Comparsion}
\vspace{-5pt}  
\end{figure*}

\subsection{Effectiveness of Perception-Aware Optimization}
\label{sec:app_perception}
We further analyze how the proposed Perception-Aware Optimization shapes the model’s behavior by observing the distribution of score differences between original and degraded images at different training stages (Steps 200/400/600), as shown in Fig.~\ref{fig:distribution_2}. The score difference of the model \textcolor[HTML]{f59161}{training without Perception-Aware Optimization} gradually concentrates near zero, indicating that the model tends to assign almost identical scores to the original and degraded images. This suggests that, without an explicit perceptual constraint, the policy mainly fits the scalar rewards and gradually loses sensitivity to quality differences between the two visual conditions.  In contrast, the score difference of the model \textcolor[HTML]{87a2ff}{training with Perception-Aware Optimization} progressively shifts toward larger gaps, indicating that the model increasingly separates its predictions for original versus degraded images as training proceeds. This behavior is consistent with the design of our perceptual KL objective, which explicitly encourages distinct output distributions under the two visual conditions. The evolving distributions confirm that Perception-Aware Optimization effectively strengthens the model’s visual perception of degradations and leads to more reliable and quality-sensitive IQA predictions.

\begin{figure*}[h]
  \centering
\includegraphics[width=0.8\linewidth]{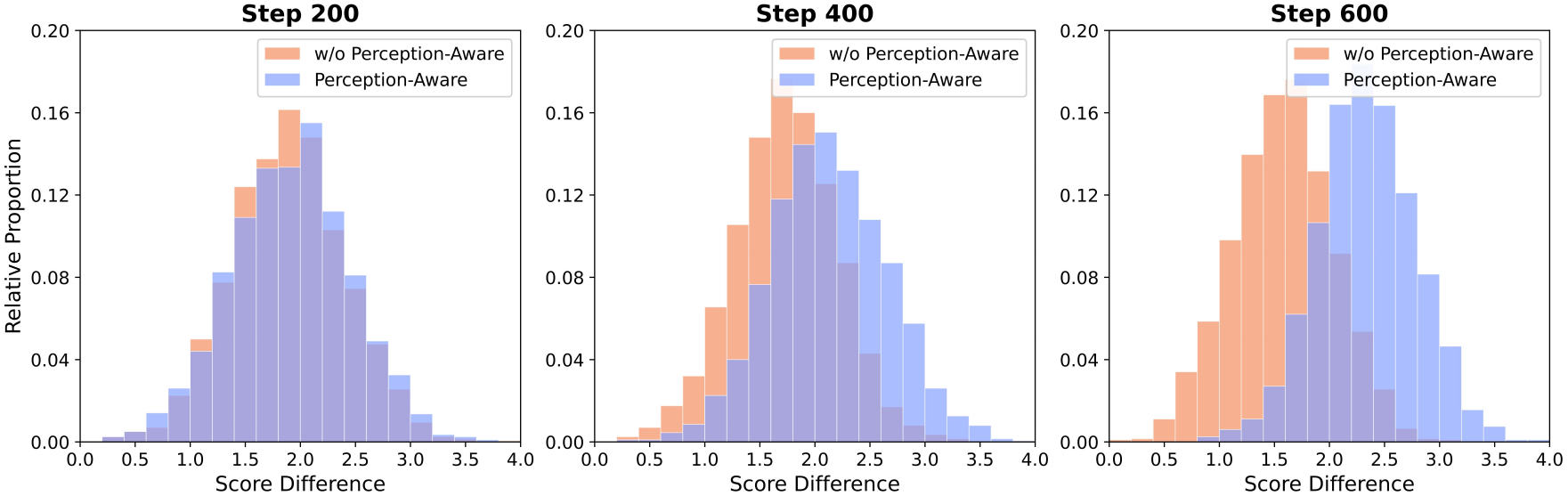}
\caption{\textbf{Analyse of effectiveness of Perception-Aware Optimization.} At different training steps, we plot histograms of score differences between original and degraded images for the model \textcolor[HTML]{F59161}{without Perception-Aware} and \textcolor[HTML]{87A2FF}{with Perception-Aware}.
Without the perceptual objective, the distribution gradually collapses toward zero, while Perception-Aware Optimization maintains and enlarges the score gap, indicating stronger sensitivity to visual quality differences.}
 \label{fig:distribution_2}
\vspace{-5pt}  
\end{figure*}

We further provide a visualization example to illustrate how Perception-Aware Optimization affects the model’s behavior, as shown in Fig.~\ref{fig:Visualization_1}. For an original--degraded image pair, the model trained \emph{without} Perception-Aware Optimization produces very similar scores and its reasoning only vaguely mentions blur or artifacts, leading to an over-optimistic assessment of the degraded image. In contrast, Q-Hawkeye with the Perception-Aware strategy explicitly attends to fine-grained visual cues such as pixelation, loss of detail, and local graininess, and accordingly assigns a noticeably lower score to the degraded image while keeping the score for the original image high and well aligned with the ground truth. These visualizations confirm that Perception-Aware Optimization guides the policy to form more faithful descriptions of degradations, thereby strengthening the reliability and quality-sensitivity of the final assessments.

\begin{figure*}[h]
  \centering
\includegraphics[width=1.0\linewidth]{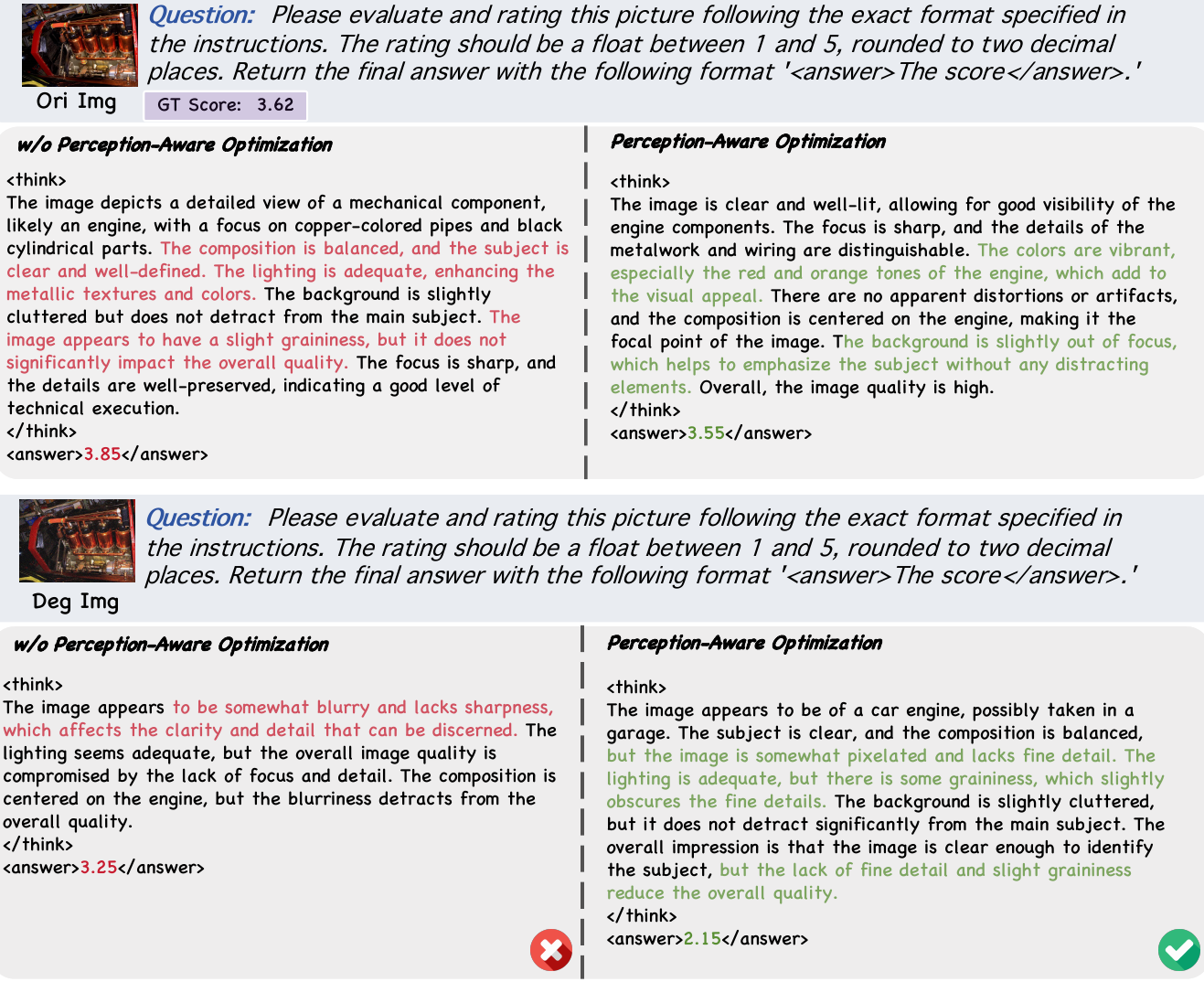}
\caption{Subjective ablation comparison between joint multi-task training and w/o joint training
on the explanation of image scoring. With joint training, our method can better perceive degradation
cues in images (such as pixelated appearance), thereby improving the accuracy of quality assessment.}
 \label{fig:Visualization_1}
\vspace{-5pt}  
\end{figure*}

\section{Complexity Analysis}
\label{sec:app_complexity}
Following~\cite{VisualQuality-R1}, we present a quantitative comparison of model complexity across representative IQA methods in Table~\ref{tab:complexity}, measured on a $512 \times 384 \times 3$ input image. Traditional CNN/Transformer-based models such as UNIQUE, MUSIQ, and MANIQA, as well as the lightweight VLM LIQE, have much fewer parameters and incur lower inference cost, but lack the reasoning and cross-dataset robustness provided by large VLMs. In contrast, Q-Align, DeQA-Score, Q-Insight, VisualQuality-R1 and our Q-Hawkeye all build on a Qwen2.5-VL-7B–scale backbone, leading to similar memory footprints during inference. Compared with other RL-based methods (Q-Insight and VisualQuality-R1), Q-Hawkeye exhibits slightly higher inference time and TFLOPs, but remains in the same order of magnitude. Overall, Q-Hawkeye achieves the best accuracy among VLM-based IQA models with only a modest overhead over existing GRPO-style baselines, showing that the gains in reliability and perceptual alignment come at a reasonable computational cost.

\begin{table}[h]
\centering
\caption{Model complexity comparison using a $512 \times 384 \times 3$ image as input.}
\label{tab:complexity}
\begin{tabular}{l|cccc}
\toprule
Method & \#Parameters & \makecell{Inference \\ Time} & \makecell{Inference \\ Memory} & \#TFLOPs \\
\midrule
UNIQUE \cite{UNIQUE} & 22.32 M & 0.02 s & 1.62 G & 0.029 \\
MUSIQ \cite{MUSIQ} & 27.13 M & 0.05 s & 1.69 G & 0.026 \\
MANIQA \cite{ManIQA} & 135.75 M & 0.03 s & 2.13 G & 0.217 \\
LIQE \cite{LIQE} & 151.28 M & 0.03 s & 2.15 G & 0.131 \\
Q-Align \cite{Q-Align} & 8.20 B & 0.14 s & 17.1 G & 1.98 \\
DeQA-Score \cite{DeQA-Score} & 8.20 B & 0.11 s & 17.1 G & 1.98 \\
Q-Insight \cite{Q-insight} & 8.29 B & 2.72 s & 17.6 G & 8.71 \\
VisualQuality-R1 \cite{Visionary-R1} & 8.29 B & 2.34 s & 17.6 G & 7.74 \\
\midrule
Q-Hawkeye (Ours) & 8.29 B & 3.10 s & 17.6 G & 9.55 \\
\bottomrule
\end{tabular}
\end{table}

\section{Detail of Evaluated Datasets}
\label{sec:app_datasets}
\begin{itemize}
\item \textbf{KonIQ}: An in-the-wild IQA dataset of 10,073 images with authentic distortions, sourced from a web photo collection, and annotated with mean opinion scores obtained via crowdsourcing (~1.2 million quality ratings from 1,459 workers).

\item \textbf{SPAQ}: A smartphone photography IQA database of 11,125 images captured by 66 mobile cameras (covering diverse real-world scenes and realistic camera distortions such as noise, blur, and exposure issues), with subjective quality scores (MOS) collected in a well-controlled laboratory study.

\item \textbf{KADID}: The KADID-10k dataset consists of 81 pristine reference images and 10,125 distorted images (25 types of synthetic distortions applied at 5 levels each), with differential mean opinion scores gathered from a crowdsourced subjective study (~30 ratings per image on average).

\item \textbf{KADIS}: The KADIS-700k dataset is a large-scale synthetic distortion IQA set containing 140,000 pristine images and 700,000 distorted images generated using 25 distortion types (five degraded versions per pristine image). Instead of subjective MOS/DMOS labels, it provides weak supervision via pre-computed scores from 11 full-reference IQA metrics for each distorted image, enabling multi-task feature learning for NR-IQA.

\item \textbf{PIPAL}: A large-scale perceptual IQA dataset containing 250 reference images and roughly 29,000 distorted images (40 distortion types spanning four subclasses: traditional distortions, super-resolution artifacts, denoising artifacts, and image blending effects). The quality labels are derived from over 1.13 million human judgments obtained using an Elo-based crowdsourcing system.

\item \textbf{LIVE-Wild}: Also known as the LIVE “In the Wild” Challenge database, it includes 1,162 images with diverse authentic distortions (photos captured via many modern mobile devices in unconstrained settings), and it was rated through a massive online crowdsourced study (yielding over 350,000 opinion scores from more than 8,100 participants).

\item \textbf{AGIQA-3K}: A novel IQA dataset of 2,982 AI-generated images produced by six different text-to-image generation models (covering GAN-based, autoregressive, and diffusion-based methods). Each image is annotated with fine-grained subjective quality scores – assessing perceptual quality and text-image alignment – collected via a controlled lab user study (reporting MOS for both dimensions).

\item \textbf{CSIQ}: A laboratory-based IQA database comprising 30 original images and 866 distorted images created by six categories of synthetic distortions (e.g., JPEG and JPEG2000 compression, Gaussian blurring, additive noise, and global contrast reduction). Subjective quality ratings (differential MOS) were obtained from 35 human observers under controlled viewing conditions.

\item \textbf{FLIVE}: The largest in-the-wild IQA dataset to date, containing 39,810 real-world images (selected from the YFCC100M social-media photo set) that exhibit a broad range of authentic distortions (variations in brightness, colorfulness, noise, sharpness, etc.). Each image is associated with a mean opinion score on a 0–100 quality scale, derived from crowdsourced human annotations.
\end{itemize}

\section{Details of Evaluation Metrics}
\label{sec:app_metrics}
Given a test set with $N$ images, let $y_i$ denote the ground-truth MOS and $\hat{y}_i$ denote the predicted quality score for the $i$-th image. In IQA, it is also common to apply a monotonic nonlinear regression (e.g., a logistic mapping) to predictions before computing correlation metrics; we denote the (optionally) mapped prediction as $\tilde{y}_i$.

\paragraph{PLCC (Pearson Linear Correlation Coefficient).}
PLCC measures the \emph{linear} agreement between predictions and MOS. It is defined as
\begin{equation}
\mathrm{PLCC}(y,\tilde{y}) =
\frac{\sum_{i=1}^{N}(y_i-\bar{y})(\tilde{y}_i-\bar{\tilde{y}})}
{\sqrt{\sum_{i=1}^{N}(y_i-\bar{y})^2}\;\sqrt{\sum_{i=1}^{N}(\tilde{y}_i-\bar{\tilde{y}})^2}},
\end{equation}
where $\bar{y}=\frac{1}{N}\sum_{i=1}^{N}y_i$ and $\bar{\tilde{y}}=\frac{1}{N}\sum_{i=1}^{N}\tilde{y}_i$.
PLCC ranges in $[-1,1]$, and a higher value indicates better linear consistency with human judgments.

\paragraph{SRCC (Spearman Rank-Order Correlation Coefficient).}
SRCC evaluates the \emph{monotonic} consistency by computing Pearson correlation on ranks. Let
$r_i=\mathrm{rank}(y_i)$ and $s_i=\mathrm{rank}(\tilde{y}_i)$ be the rank indices (with average ranks for ties). SRCC is computed as
\begin{equation}
\mathrm{SRCC}(y,\tilde{y}) =
\frac{\sum_{i=1}^{N}(r_i-\bar{r})(s_i-\bar{s})}
{\sqrt{\sum_{i=1}^{N}(r_i-\bar{r})^2}\;\sqrt{\sum_{i=1}^{N}(s_i-\bar{s})^2}},
\end{equation}
where $\bar{r}=\frac{1}{N}\sum_{i=1}^{N}r_i$ and $\bar{s}=\frac{1}{N}\sum_{i=1}^{N}s_i$.
When there are no tied ranks, SRCC can be equivalently written as
\begin{equation}
\mathrm{SRCC}(y,\tilde{y}) = 1 - \frac{6\sum_{i=1}^{N} d_i^2}{N(N^2-1)}, \quad d_i = r_i - s_i.
\end{equation}
Similar to PLCC, SRCC lies in $[-1,1]$, and a larger value indicates better rank-order agreement.

\section{Details of Degraded Image Construction}
\label{sec:app_degraded}
To explicitly train Q-Hawkeye to perceive visual quality differences, we construct a paired origin--degraded training set by applying four types of degradations to each original image $I$: Noise, Blur, JPEG, and Darken. For each sample, one degradation type is randomly sampled and applied with fixed hyperparameters, while keeping the task prompt $q$ unchanged. The concrete operators and parameter settings are summarized in Tab.~\ref{tab:deg_params}. 

Concretely, the Noise branch adds zero-mean additive Gaussian noise on the RGB channels, controlled by the noise standard deviation $\sigma_{\text{noise}}$. The Blur branch applies an isotropic Gaussian blur with standard deviation $\sigma_{\text{blur}}$, simulating out-of-focus or motion blur. The JPEG branch compresses the image using a JPEG encoder with a specified quality factor $q$, introducing typical blocking and ringing artifacts. The Darken branch scales the pixel intensities by a factor $\lambda$, reducing global brightness and contrast while keeping the content unchanged. 

Some origin--degraded pairs may still be too weak to induce a noticeable perceptual change (e.g., applying additional blur to an already blurred image), which fails to provide an effective contrastive signal. To ensure that each retained pair $(I, I^{\mathrm{deg}})$ exhibits a clear and consistent perceptual difference, we enforce the ``effective contrast'' requirement using the double-check filter strategy described in the main paper. Specifically, we first employ a strong VLM (GPT-4o) to judge whether $(I, I^{\mathrm{deg}})$ contains a discernible visual difference. For pairs that are flagged as indistinguishable, we further verify them with human experts. If a pair is ultimately confirmed to have no discernible quality difference, we resample the degradation type and regenerate $I^{\mathrm{deg}}$. We repeat this process until all retained pairs can be consistently distinguished. On the KonIQ training split (approximately 6.8K images), after the first-stage VLM screening, about 610 degraded images are flagged as indistinguishable from their corresponding originals. After the second-stage human verification, 31 image pairs are confirmed to have no discernible quality difference, and are thus regenerated via resampling to satisfy the ``effective contrast'' requirement.  This construction provides diverse yet controllable degradations and supplies Q-Hawkeye with reliable supervision to learn fine-grained visual sensitivity to quality changes.

\begin{table}[t]
\centering
\caption{\textbf{Degradation configurations} used to construct original--degraded training pairs.}
\label{tab:deg_params}
\resizebox{0.6\linewidth}{!}{%
\begin{tabular}{l l c c}
\toprule
\textbf{Type} & \textbf{Operation} & \textbf{Parameter} & \textbf{Value used} \\
\midrule
Noise   & Additive Gaussian noise
        & $\sigma_{\text{noise}}$
        & $45$ \\
Blur    & Gaussian blur
        & $r_{\text{blur}}$
        & $2.0$ \\
JPEG    & Repeated JPEG compression
        & $Q_{\text{JPEG}}$
        & $5$ \\
Darken  & Brightness reduction
        & $\lambda_{\text{dark}}$
        & $0.6$ \\
\bottomrule
\end{tabular}%
}
\end{table}

\section{Details of Hyperparameters}
\label{app:hyperparameters}

Our objective (Eq.~\ref{eq:total_objective}) introduces a small set of hyperparameters. The uncertainty temperature $\tau$ controls the strength of uncertainty-aware reweighting $w=\exp(-\tau\tilde u)$, which determines how aggressively high-uncertainty samples are downweighted to prevent noisy gradients from unstable rollouts from dominating GRPO updates. The perception weight $\gamma$ governs the implicit perception constraint that maximizes the distributional discrepancy between pristine and degraded inputs, enforcing that quality judgments are grounded in genuine visual evidence. However, maximizing this discrepancy alone can be exploited by increasing output randomness, so we introduce double entropy regularization with weights $\eta_1,\eta_2$ to prevent high-entropy collapse and stabilize training; we tie $\eta_1\!=\!\eta_2$ to reduce degrees of freedom while retaining the stabilizing effect. All remaining hyperparameters (e.g., $K$ rollouts, reward tolerance $\alpha$, and GRPO KL weight $\beta$) follow standard RL post-training practice and are kept fixed across datasets; importantly, we use a single default configuration (e.g., $\tau\!=\!0.2$, $\gamma\!=\!5\times10^{-4}$, $\eta_1\!=\!\eta_2\!=\!1\times10^{-4}$) without per-dataset tuning in all main experiments.

To further verify the robustness of Q-Hawkeye with respect to hyperparameter choices, we conduct extensive \emph{joint} sensitivity studies over the key hyperparameters and evaluate on all eight benchmarks (PLCC values reported in the heatmaps). Fig.~\ref{fig:gamma_eta} sweeps $\gamma\in\{10^{-3},5\times10^{-4},1\times10^{-4}\}$ against $\eta_1\!=\!\eta_2\in\{10^{-3},5\times10^{-4},1\times10^{-4}\}$ (with $\tau$ fixed), showing a broad high-performance plateau where the average performance stays within a stable and strong regime across the full grid. Fig.~\ref{fig:gamma_tau} further sweeps $\gamma$ against $\tau\in\{0.2,0.3,0.4\}$ (with $\eta_1\!=\!\eta_2$ fixed), again exhibiting stable behavior over a reasonable range: for moderate uncertainty suppression (e.g., $\tau\in\{0.2,0.3\}$), the overall performance remains consistently strong and varies smoothly with $\gamma$; when $\tau$ becomes overly large (e.g., $\tau\!=\!0.4$), performance degrades in a predictable and interpretable manner, consistent with the intuition that excessive uncertainty suppression may lead to under-updating rather than training instability. Overall, these experiments demonstrate that our method admits a wide operating region with consistently strong performance, supporting the robustness of the proposed hyperparameters.

\begin{figure*}[h]
  \centering
\includegraphics[width=0.5\linewidth]{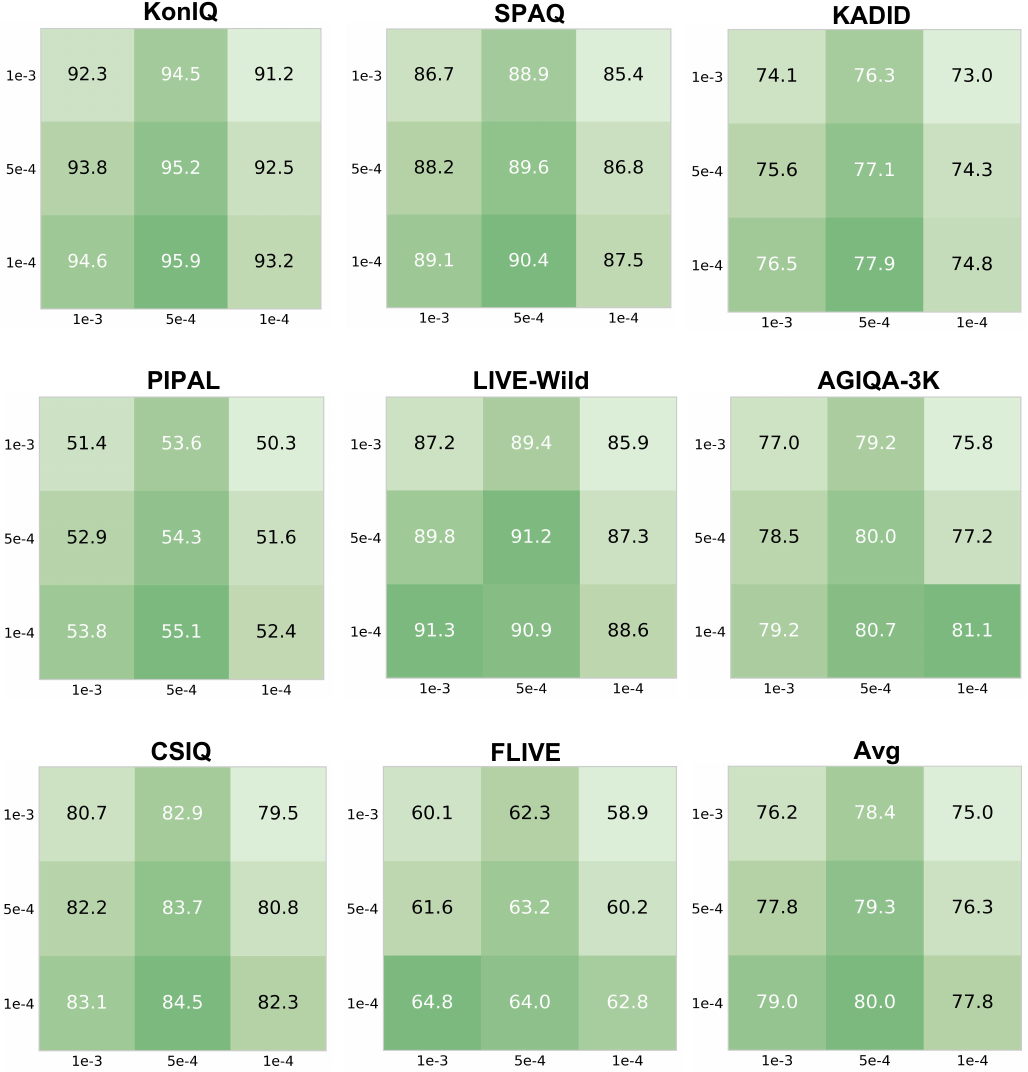}
\caption{Sensitivity experiments on $\gamma$ and $\tau$ (with $\eta_1\!=\!\eta_2$ fixed), evaluated on all eight benchmarks (PLCC$\times 100$). Performance remains stable for moderate $\tau$.}
 \label{fig:gamma_eta}
\end{figure*}

\begin{figure*}[h]
  \centering
\includegraphics[width=0.5\linewidth]{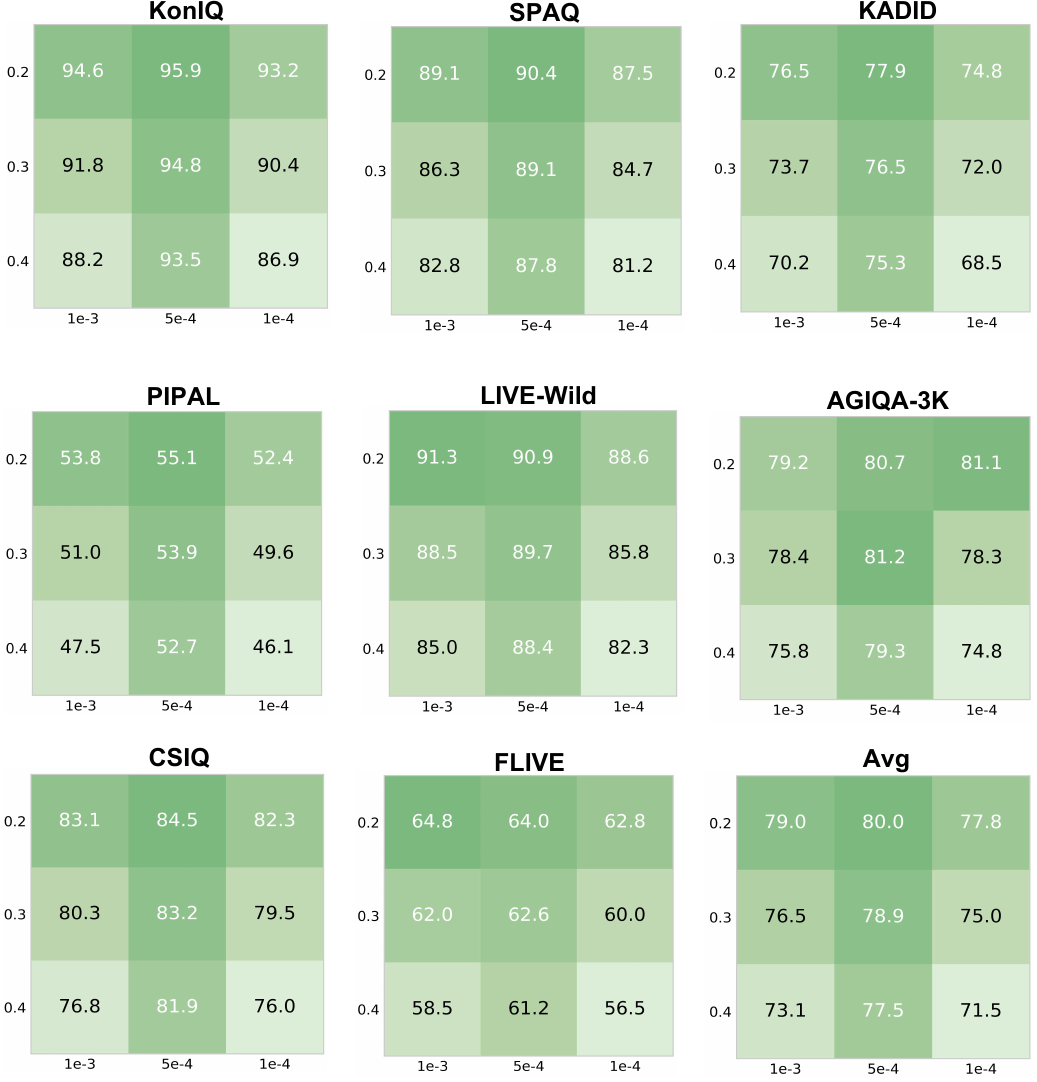}
\caption{This figure shows the performance sensitivity of Q-Hawkeye with respect to the hyperparameters $\gamma$ and $\eta_1 = \eta_2$. The heatmap demonstrates the robustness of our method, where performance remains stable within a strong range for various combinations of these hyperparameters, maintaining consistent results across multiple datasets.}
 \label{fig:gamma_tau}
\end{figure*}

\section{More Visualization Examples}
\label{sec:app_vis}
We provide more visualization examples of quality assessment under different degradation conditions in Fig.~\ref{fig:Visualization_2}-\ref{fig:Visualization_5} to further evaluate Q-Hawkeye's visual perception and assessment capabilities under different visual conditions. As can be seen, although Q-Hawkeye is trained using only the KONIQ training set, compared to methods like Q-insight and VisualQuality-R1 that use multiple training sets for joint training, Q-Hawkeye can perceive perceptual differences across different visions and provide more accurate assessment scores.

\begin{figure*}[h]
  \centering
\includegraphics[width=1.0\linewidth]{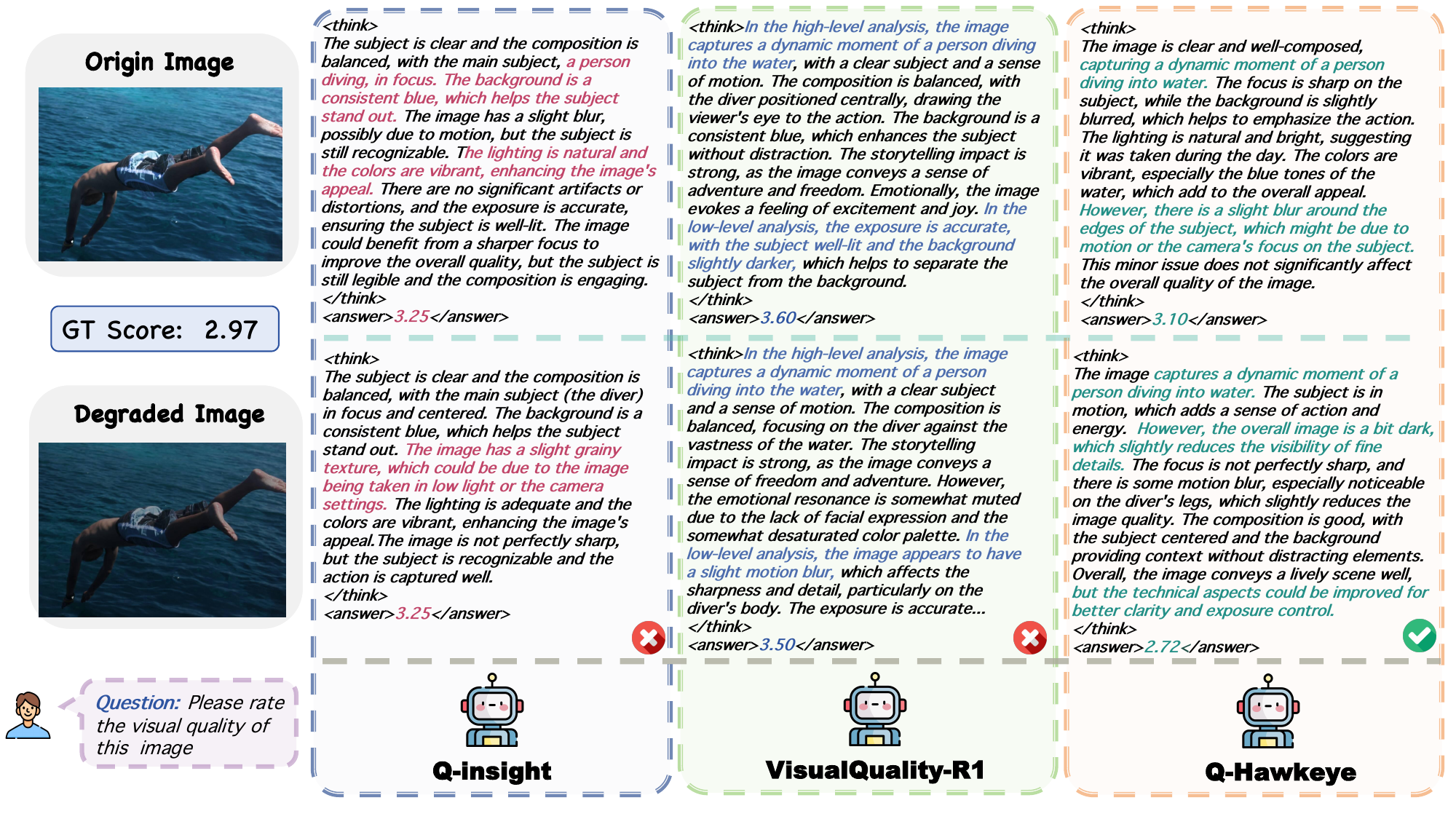}
\caption{Visual examples of quality assessment for three methods under \textbf{Darken} degradation conditions}
 \label{fig:Visualization_2}
\vspace{-5pt}  
\end{figure*}

\begin{figure*}[h]
  \centering
\includegraphics[width=1.0\linewidth]{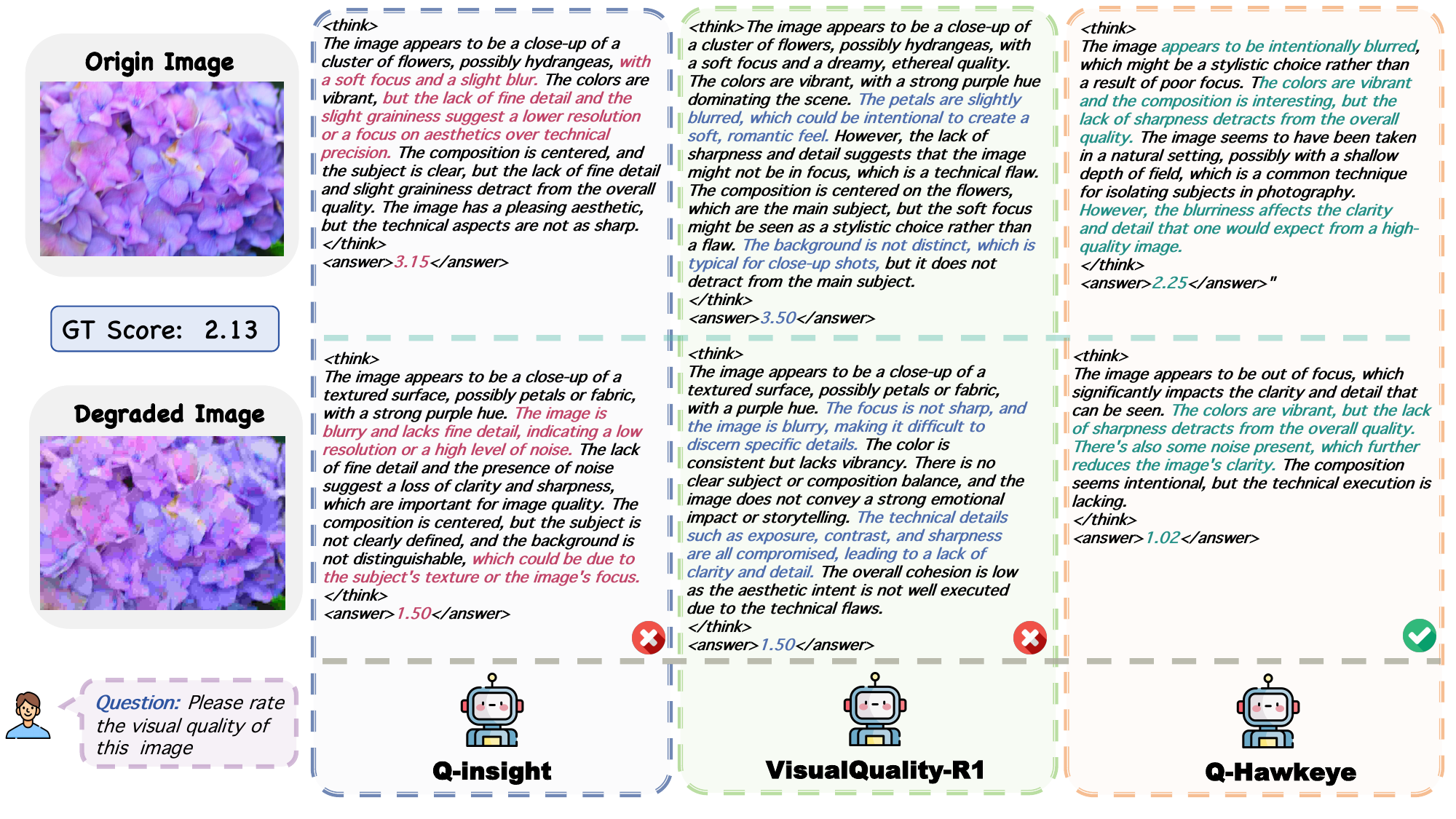}
\caption{Visual examples of quality assessment for three methods under \textbf{JPEG} degradation conditions}
 \label{fig:Visualization_3}
\vspace{-5pt}  
\end{figure*}

\begin{figure*}[h]
  \centering
\includegraphics[width=1.0\linewidth]{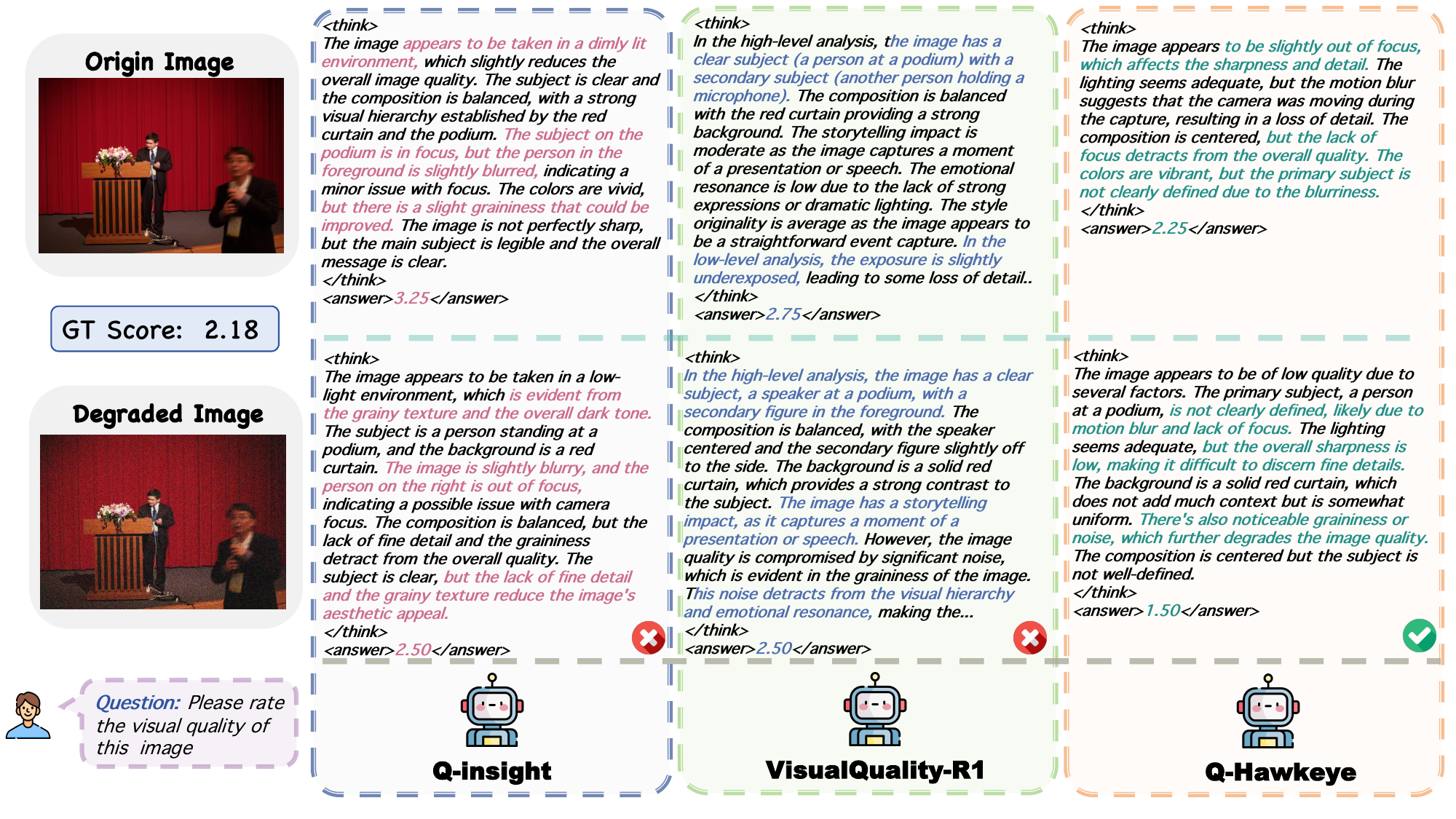}
\caption{Visual examples of quality assessment for three methods under \textbf{Noise} degradation conditions}
 \label{fig:Visualization_4}
\vspace{-5pt}  
\end{figure*}

\begin{figure*}[h]
  \centering
\includegraphics[width=1.0\linewidth]{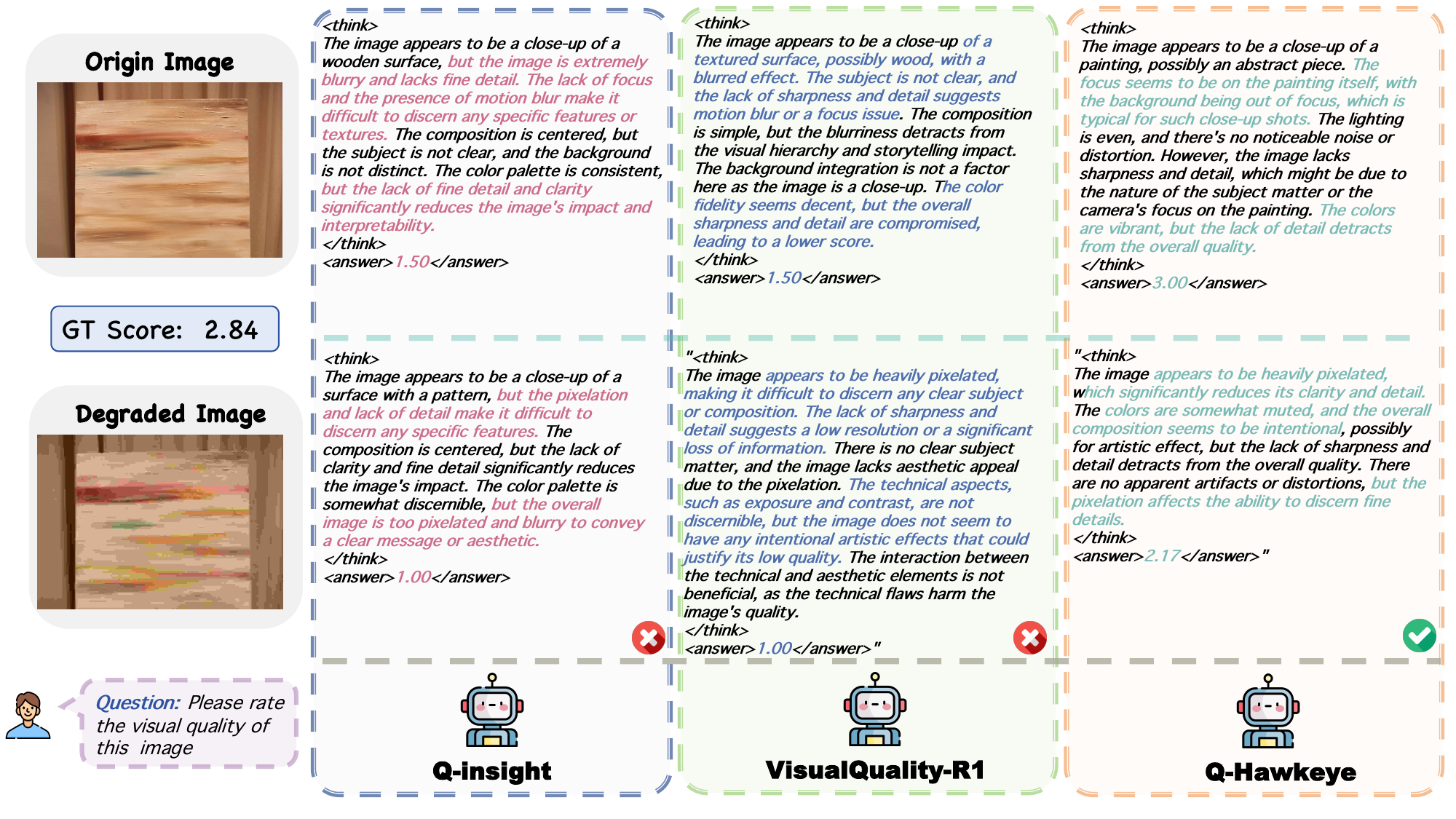}
\caption{Visual examples of quality assessment for three methods under \textbf{Blur} degradation conditions}
 \label{fig:Visualization_5}
\vspace{-5pt}  
\end{figure*}

\begin{figure*}[h]
  \centering
\includegraphics[width=0.7\linewidth]{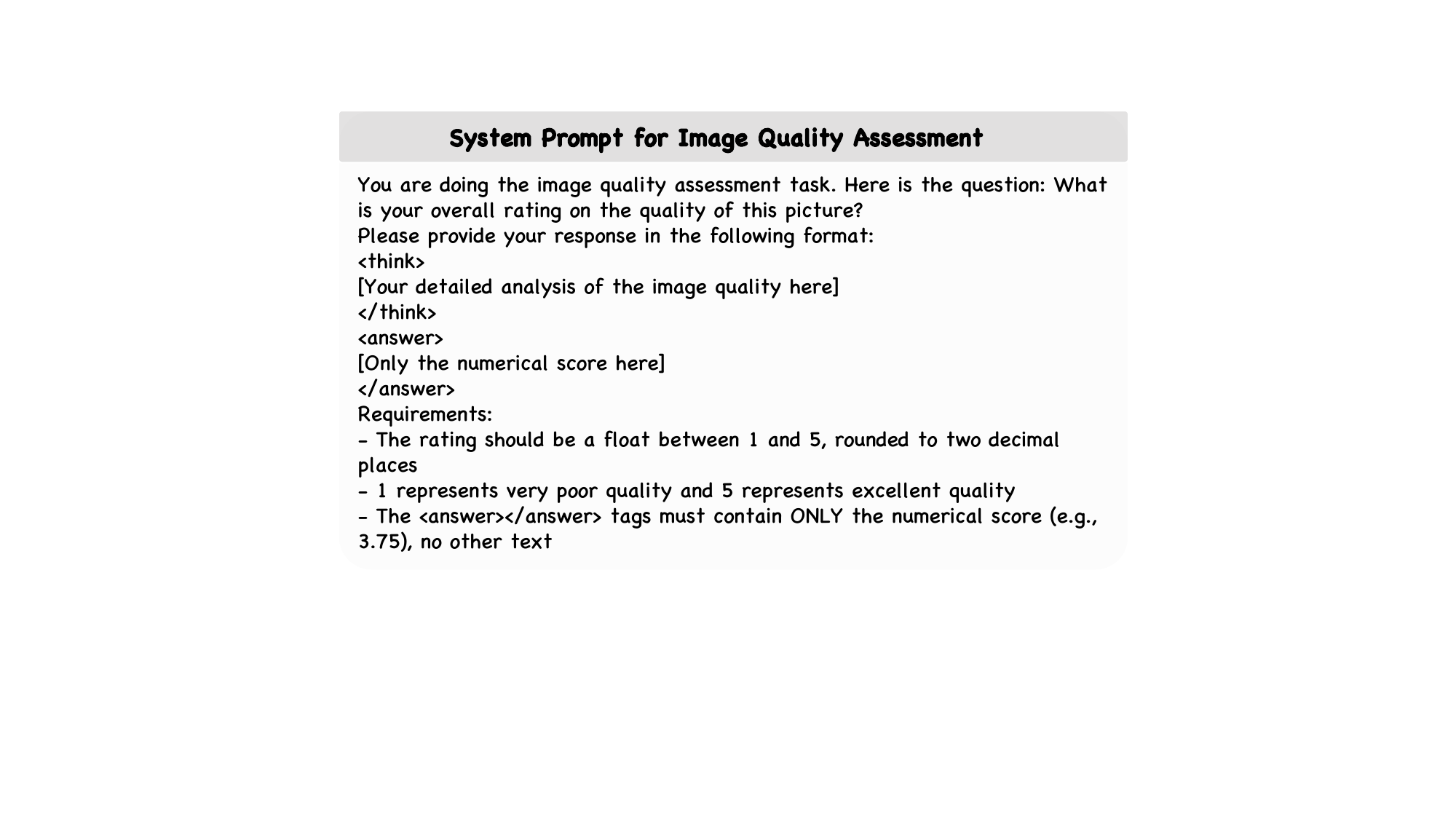}
\caption{System Prompt for Image Quality Assessment task.}
 \label{fig:SystemPrompt}
\vspace{-5pt}  
\end{figure*}

\begin{algorithm}[h]
\caption{Q-Hawkeye}
\label{alg:qhawkeye}
\textbf{Input:} IQA dataset $\mathcal{D} = \{(I,q,y)\}$; degraded pair set
$\mathcal{D}^{\mathrm{deg}} = \{(I,I^{\mathrm{deg}},q)\}$; initial policy $\pi_\theta$;
reference policy $\pi_{\mathrm{ref}}$; rollouts per input $K$; batch size $B$; epochs $E$.\\
\textbf{Output:} Updated policy $\pi_{\theta}^{\phi }$.

\begin{algorithmic}[1]
\FOR{$e \gets 1$ \textbf{to} $E$}
  \FOR{each mini-batch $\mathcal{B} \subset \mathcal{D}$ of size $B$}
    \STATE \textit{// Uncertainty-Aware Dynamic Optimization}
    \FORALL{$(I,q,y) \in \mathcal{B}$}
      \STATE Sample $K$ rollouts $o_k \sim \pi_{\theta_{\mathrm{old}}}(\cdot \mid I,q)$
      \STATE Compute rewards $r_k$ by Eqs.~(1)--(3)
      \STATE Compute advantages $A_k$ by Eq.~(5)
      \STATE Estimate uncertainty $u$, $\tilde{u}$, $w(u)$ by Eqs.~(12)--(15)
      \STATE Get $\tilde{A}_k$ by Eq.~(16)
    \ENDFOR
    \STATE \textit{// Perception-Aware Optimization}
    \FORALL{$(I,q,y) \in \mathcal{B}$}
      \STATE Retrieve $I^{\mathrm{deg}}$ from $\mathcal{D}^{\mathrm{deg}}$
      \STATE Compute $D_{\mathrm{IPL}}$ by Eqs.~(17)--(19)
      \STATE Compute $\mathcal{L}_{\mathrm{ent}}$ by Eqs.~(20)--(21)
    \ENDFOR
    \STATE Update $\theta$ using the overall objective $\mathcal{L}_{\mathrm{Total}}$ by Eq.~(22)
  \ENDFOR
\ENDFOR
\STATE \textbf{return} $\pi_{\theta}^{\phi }$
\end{algorithmic}
\end{algorithm}

\end{document}